\documentclass{article}

\usepackage{hyperref}

\usepackage[accepted]{mlsys2022}
\usepackage[utf8]{inputenc}
\usepackage{textcomp}
\usepackage{microtype}
\usepackage{graphicx}
\usepackage{subfigure}
\usepackage{booktabs} 
\usepackage{soul}
\usepackage{enumitem}
\usepackage{amsmath}
\usepackage{makecell}
\usepackage{multirow}
\usepackage{vwcol}
\usepackage{siunitx}
\usepackage{fdsymbol}
\usepackage{xfrac}
\usepackage[table]{xcolor}
\usepackage{algpseudocode} 
\definecolor{lightergray}{gray}{0.93}

\DeclareUnicodeCharacter{03BC}{\TextOrMath{\textmu}{\mu}}

\setitemize{topsep=0pt, itemsep=0pt, parsep=2pt}
\setenumerate{topsep=0pt, itemsep=0pt, parsep=2pt}

\mlsystitlerunning{Differentiable Network Pruning for Microcontrollers}

\makeatletter
\renewcommand{\mlsys@appearing}{
Author's version. Full version available in Proceedings of the ACM on Interactive, Mobile, Wearable and Ubiquitous Technologies, Vol. 6, No. 4 \url{https://doi.org/10.1145/3569468}
}
\makeatother

\begin{document}

\twocolumn[
\mlsystitle{Differentiable Network Pruning for Microcontrollers}




\begin{mlsysauthorlist}
\mlsysauthor{Edgar Liberis}{cam,sam}
\mlsysauthor{Nicholas D. Lane}{cam,sam}
\end{mlsysauthorlist}

\mlsysaffiliation{cam}{Department of Computer Science and Technology, University of Cambridge, UK}
\mlsysaffiliation{sam}{Samsung AI Centre Cambridge, UK}

\mlsyscorrespondingauthor{Edgar Liberis}{el398 \textit{at} cam.ac.uk}

\mlsyskeywords{Machine Learning, MLSys, MCU, NAS, Deep learning, Neural Networks, Model Compression, Architecture Search, Microcontrollers, TinyML}

\vskip 0.3in

\begin{abstract}
Embedded and personal IoT devices are powered by microcontroller units (MCUs), whose extreme resource scarcity is a major obstacle for applications relying on on-device deep learning inference. Orders of magnitude less storage, memory and computational capacity, compared to what is typically required to execute neural networks, impose strict structural constraints on the network architecture and call for specialist model compression methodology. In this work, we present a differentiable structured network pruning method for convolutional neural networks, which integrates a model's MCU-specific resource usage and parameter importance feedback to obtain highly compressed yet accurate classification models. Our methodology (a) improves key resource usage of models up to 80$\times$; (b) prunes iteratively while a model is trained, resulting in little to no overhead or even improved training time; (c) produces compressed models with matching or improved resource usage up to 1.4$\times$ in less time compared to prior MCU-specific methods. Compressed models are available for download.\footnotemark
\end{abstract}
]



\printAffiliationsAndNotice{}  

\section{Introduction}
\label{introduction}


\begin{table*}[t]
\centering
\small
\begin{tabular}{lrrrrrr}
\toprule
\textbf{Device} & \textbf{Cores} & \textbf{Freq. (GHz)} & \textbf{RAM (GB)} & \textbf{Storage (GB)} & \textbf{Power (W)} & \textbf{Price (\$)} \\
\hline
NVIDIA A100 GPU & 6912 & 1.4 & 40 & --- & 400 & 12,500 \\
Galaxy S22 Smartphone & 8 & $<$ 2.8 & 8 & 128 & $\approx$3.6 & 750 \\
Raspberry Pi 3B+ & 4 & 1.4 & 1 & 32--512 & 2.3 & 40 \\
\href{https://os.mbed.com/platforms/ST-Nucleo-F767ZI/}{NUCLEO-F767ZI} (M7) & 1 & 0.216 & 0.000512 & 0.002 & 0.3 & 9 \\
\href{https://os.mbed.com/platforms/ST-Nucleo-F446RE/}{NUCLEO-F446RE} (M4) & 1 & 0.180 & 0.000128 & 0.000512 & 0.1 & 3 \\
\bottomrule
\end{tabular}
\label{tbl:devices-comp}
\caption{Specifications of a high-end GPU server, a smartphone, a micro-computer and Nucleo development boards with ARM's Cortex M7 and M4 microcontrollers. Microcontrollers are the most resource-constrained yet the most power-efficient and cheapest platform.}
\end{table*}

\footnotetext{\url{https://bit.ly/3FQNmws}}

There's an increasing interest in bringing on-device neural networks to what is arguably the smallest-scale viable compute platform: microcontroller-powered IoT devices. The roadblock to this, however, is the limited compute power. Microcontroller units (MCUs) are single-chip systems, typically consisting of a single-core processor (such as ARM M-series), on-chip persistent Flash and temporary SRAM memories, other peripherals (e.g. sensors, microphones, radio).  Table~\ref{tbl:devices-comp} compares technical specifications of MCUs against data center, mobile and micro-computer hardware. MCUs are designed with cost and power-efficiency in mind, which is primarily achieved by reducing the on-board memory and compute resources. The data shows that, even when compared to micro-computers like Raspberry Pi, MCUs have orders of magnitude less storage and SRAM (GBs vs 100s KB) and a slower processor (GHz vs 100s MHz). 

\begin{figure}[t]
    \centering
    \includegraphics[scale=1.25]{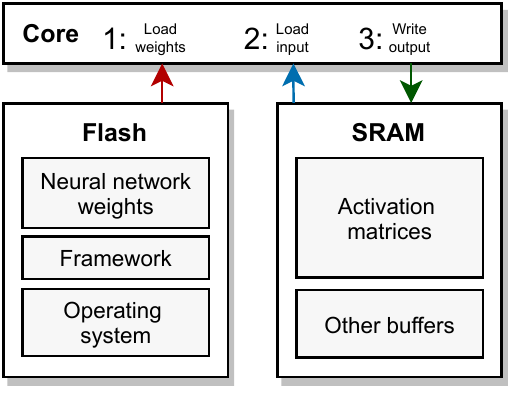}%
    \caption{Deep learning inference on an MCU illustrated using relevant components. A flat memory hierarchy and its limited capacity impose significant constraints on the network architecture.}
    \label{fig:mcu-comp}
\end{figure}

For many practitioners, it is not trivial to work around the resource scarcity of microcontrollers. As a result, applications choose to offload the execution of neural networks to a more capable device, such as a smartphone or a remote ``cloud'' backend server. This brings numerous disadvantages, such as a lowered degree of privacy, complete reliance on a working communication link, increased response latency, and increased power usage for radio communication.

To see what properties the resource scarcity imposes on neural networks, let us take a more detailed look at how a neural network's execution may be mapped to a microcontroller (Figure~\ref{fig:mcu-comp}), such as via the TensorFlow Lite Micro runtime~\cite{david2020tensorflow}:
\begin{itemize}
    \item Each layer of a neural network will be executed in some predefined order, one at a time and in its entirety.
    \item A layer is executed by loading its parameters (weights) from Flash, loading its input from SRAM and writing the output back to SRAM.
    \item All constant data, such as weights of a neural network and program code, are stored in the read-only Flash memory. Therefore, the size of a neural network is limited by Flash (e.g. $\leq$ 64KB).
    \item All temporary intermediate data, including activation matrices, will be stored in SRAM. Therefore, a neural network's peak activation size / memory usage is limited by the amount of SRAM (e.g. $\leq$ 64 KB).
\end{itemize}

As a compute platform for deep learning, microcontrollers pose unique challenges that do not similarly manifest on other platforms. The need to constrain SRAM usage, a relatively slow processor and orders of magnitude fewer resources requires \emph{further specialist research} than what is typically explored in GPU/mobile-scale neural network design and compression. In fact, the need for MCU-specific methodology has spawned a research subfield called \emph{TinyML}, which includes runtime design~\cite{david2020tensorflow, liberis2019neural}, optimised layer implementations~\cite{lai2018cmsis} and task-specific MCU-sized model discovery~\cite{fedorov2019sparse, liberis2021munas, banbury2020micronets}.
 
\begin{figure*}[t]
    \centering
    \includegraphics[scale=0.98]{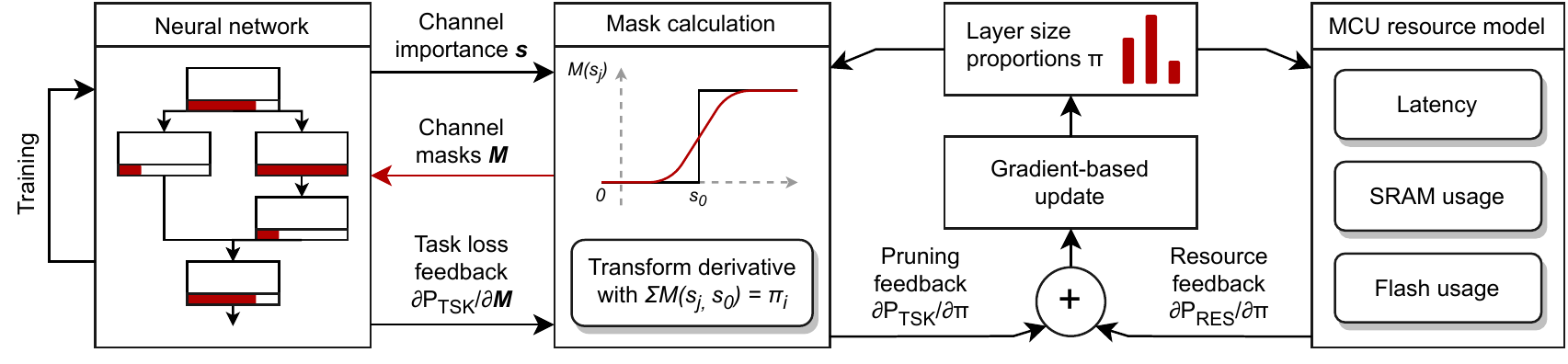}%
    \caption{A graphical summary of the proposed method. Both task-specific and resource usage-specific objectives influence pruning.}
    \label{fig:overall-diag}
\end{figure*}

In this work (Figure~\ref{fig:overall-diag}), we tackle this problem through the use of \emph{budgeted differentiable network pruning}. We employ bi-level gradient descent optimisation to learn the sizes of each layer in the network, resulting in iterative compression that is interleaved with network training and has negligible overhead. Our methodology formulates model size, peak memory usage and latency constraints as resource budget requirements and incorporates them into pruning as differentiable objectives. This achieves MCU specialisation currently absent from pruning literature and, compared to other MCU-level compression methods, the use of pruning and accurate objectives achieves faster compression while correctly targeting resource bottlenecks.

Differentiable pruning is evaluated on benchmark audio and image classification tasks, suitable for microcontrollers, and similar to real-world sensing applications: Speech Commands keyword spotting~\cite{warden2018speech}, Visual Wake Words~\cite{chowdhery2019visual} person detection, CIFAR-10~\cite{krizhevsky09learning} and ImageNet~\cite{deng2009imagenet} image classification. Additionally, a range of investigative studies is carried out to establish that: (a) the resource objectives are required for finding microcontroller-compatible models; (b) an accurate minimal peak memory usage computation is essential for correctly allocating the SRAM budget; (c) pruning allocates resources better than popular manual uniform scaling on MobileNet-v2~\cite{sandler2018mobilenetv2}; (d) training can be accelerated by terminating pruning early compared to training without pruning; (e) pruning prioritises computationally-expensive layers of the network; (f) pruned models have viable latency and energy consumption on a microcontroller to be deployed in practice.

In summary, this work contributes to microcontroller-aware model compression by:
\begin{enumerate}
    \item \emph{Proposing a structured differentiable pruning method} for finding microcontroller-compatible models. Compared to related microcontroller-specialist work, the method accurately targets resource usage for neural networks and offers fast compression with negligible overhead during training (or even accelerates it).
    \item \emph{Conducting ablation, comparative and analysis studies} to show that each aspect of this methodology is essential for achieving viable and performant microcontroller-compatible models.
    \item \emph{Exercising differentiable pruning on {fourteen} architecture and dataset combinations} to produce state-of-the-art models at a variety of microcontroller-friendly resource usage \emph{vs} accuracy trade-off points. Results show an improved key resource usage of up to 40\% compared to related work and up to 80$\times$ compared to original models.
\end{enumerate}

\section{Related work}
\label{sec:related-work}

This section will describe how our methodology fits within the wider taxonomy of pruning algorithms and compare it to existing methods for MCU deep learning.

\subsection{Network pruning}

Network pruning removes parameters from a network that are deemed the least important for generalisation. This inherently introduces a trade-off between the predictive performance (here, classification accuracy) and the resource footprint of a network. 
Perhaps surprisingly, having conducted a wide literature survey of 81 pruning techniques, \citet{blalock2020state} cannot find significant chronological trends in improving this trade-off. 

What is, therefore, the utility of yet another pruning method? Despite comparable gains in compression rates, pruning methodologies exhibit a variety of secondary characteristics. These include: (a) computational overhead or cost of pruning; (b) the ability to produce sparse or dense models; (c) the amount of additional training required to recover from parameter loss; (d) the ability to optimise metrics, defined as a function of the network architecture; (e) ease of use and hyperparameter sensitivity; (f) over-regularisation of surviving parameters of the network. The characteristics needed for microcontroller specialisation or the usability of this methodology can be selected by carefully considering these (un-)desirable properties during the design stage. 

We classify our methodology as \emph{structured}, \emph{during-training}, \emph{iterative}, \emph{budgeted} and \emph{differentiable} pruning, with each category explained below. 

\textbf{Structured \emph{vs} unstructured pruning.} Pruning is often used to obtain sparse models, which require sparse matrix multiplication implementations or hardware support to execute efficiently~\cite{anwar2015structured}. This work implements structured pruning, which effectively performs hyperparameter adjustment by removing entire feature maps. This results in a dense (not sparse) model that can be efficiently executed using widely-supported microcontroller deep learning software. 

\textbf{Budgeted \emph{vs} ``opportunistic'' pruning.}
Resource usage-unaware pruning methods typically use the total sparsity (the proportion of feature maps or channels removed) in the model across all layers to describe the extent of model compression. In practice, this is not a descriptive metric that can be used to target a particular resource budget because the same sparsity level can result in a range of model sizes and MAC operations. A popular way to achieve resource awareness is to add a sparsifying group $L_1$ regularisation~\cite{gordon2018morphnet, liu2017learning} to all parameters of the network, followed by tuning its strength until the desired resource usage is reached. Purpose-made budgeted pruning~\cite{lemaire2019structured, ning2020dsa}, such as this work, allocates resources outside of the training loss and, therefore, does not rely on remaining channels to generalise despite additional regularisation. 

\textbf{Pre-, during- and post-training pruning.}
Post-training pruning removes parameters from a converged model, often followed by fine-tuning to recover any lost accuracy~\cite{han2015learning}. Fine-tuning and training can be combined by performing pruning during training, saving time~\cite{lin2020dynamic}. Pre-training methods use initialisation and gradient-flow information~\cite{wang2020picking} to decide what to prune before training or very early on. Performing pruning as early as possible avoids the computational overhead associated with extraneous training and enables the compressed model to be used for remaining training, reducing the overall training time.

\textbf{Differentiable \emph{vs} non-differentiable pruning.}
An essential consideration in pruning is the distribution of sparsity across network layers. Per-layer sparsity can be decided by a separate optimisation process, which can support custom resource usage metrics expressed as functions of layer width. The introduction of an architectural optimisation process brings concerns previously seen in NAS, as the optimisation needs efficient estimates of the network classification accuracy (\emph{e.g.} a proxy model~\cite{liu2019autocompress}) to push back against resource usage objectives. In this instance, differentiable optimisation methods allow for efficient iterative bi-level optimisation---both pruning (learning layer sizes) and training are performed using gradient descent~\cite{ning2020dsa}---that improves upon time-consuming alternative search algorithms like evolutionary search.

\textbf{Iterative \emph{vs} single-shot pruning.} Compared to removing all required parameters at once, repeatedly removing a small number of parameters was shown to give better recovery from parameter loss at high pruning ratios~\cite{han2015learning}. Differentiable pruning is performed iteratively, which allows this recovery to happen during training and not at a subsequent fine-tuning stage.

Among prior network pruning methodologies, Differentiable Sparsity Allocation (DSA)~\cite{ning2020dsa} is the closest work. The methodologies share the same insight to patch through classification loss gradients from masks to learnable channel width multipliers (see Section~\ref{sec:methodology}). Otherwise, methodologies differ in design and implementation details: DSA uses slow ADMM optimisation and lacks microcontroller-specific optimisation objectives.

\subsection{Deep learning for microcontrollers}

Designing MCU-compatible neural network architectures is a challenging task due to the need to obtain a sufficiently high predictive performance while maintaining a sufficiently small resource footprint. Existing MCU-specific deep learning methods that achieve this can be classified into (a) model compression methods applied to manually designed networks, such as weight quantisation or binarisation~\cite{zhang2017hello, mocerino2019coopnet}, pruning (this work) or channel search~\cite{banbury2020micronets}; (b) neural architecture search (NAS)~\cite{lin2020mcunet}. Search methods can leverage simple pruning as a nested step to further reduce the resource footprint of a candidate network~\cite{liberis2021munas, fedorov2019sparse}. 

In the extreme, NAS methods may only search for layer width hyperparameters to maximise search efficiency, which is, traditionally, the territory of structured pruning.
As a result, the closest microcontroller specialist work to ours is the ``\emph{MicroNets}''~\cite{banbury2020micronets} NAS system, which finds layer sizes for particular task-specific architectures using differentiable optimisation. Provided with a backbone architecture, the method works by masking each layer at a predefined set of widths ($10\%, 20\%, \ldots, 100\%$)---eventually, only one will be included in the final architecture, followed by fine-tuning. Experimentally, we find that both methods can discover models with similar classification accuracy and resource usage given the same backbone architecture. However, differentiable pruning offers additional benefits in comparison: (a) the pruning is guided by an improved peak memory usage calculation which is more accurate for network architectures with branches and parallel paths; (b) improved running time: compared to architecture search, pruning has negligible overhead, requires no fine-tuning and can even reduce the amount of computation needed for training by switching to a pruned model after layer widths have been established; (c) pruning allocates the resource budget more efficiently by operating at a per-channel granularity, not in increments of 10\%.

\section{Differentiable pruning for MCUs}
\label{sec:methodology}

Here, we describe (a) the neural network resource usage calculation for MCUs; (b) the task-specific pruning objective; (c) how the two are combined into a differentiable function of a network's layer sizes; (d) how the latter is used to create a differentiable budgeted pruning algorithm. The unique addition of MCU-specific resource usage to pruning allows for the fast discovery of MCU-friendly models. 

\subsection{Execution constraints}
\label{sec:exec-constr}

We formulate model latency, peak memory usage and size constraints that are both differentiable and faithful to the real resource usage of a neural network to guide pruning towards MCU-compatible models.  

\textbf{Model latency.} To ensure that the pruning algorithm creates models that run sufficiently quickly, we implement a model latency constraint. For GPU-powered systems, model latency can be either directly measured or predicted using sophisticated models.~\cite{chau2020brp} However, on MCU-powered systems, model latency can be predicted using simple metrics, such as the number of floating-point operations (FLOPs) or multiply-accumulate operations (MACs), due to the lack of any performance-enhancing features, like parallelism or caching~\cite{liberis2021munas}. Latency has also been shown to be a good predictor of power consumption on MCUs~\cite{banbury2020micronets}, due to the processor staying in the highest power state throughout the inference.

\textbf{Peak memory usage.} Peak memory usage at inference is an uncommon resource property to target in model compression. Most consumer hardware, such as devices presented in Table~\ref{tbl:devices-comp}, have a multi-level memory hierarchy, with each layer offering more capacity at a slower access speed: multiple levels of fast but small SRAM data and instruction caches co-located with the CPU, followed by at least 1 GB of RAM. The latter is sufficient for most compressed neural networks. MCUs, on the other hand, lack a memory that resides on the same capacity \textit{vs} access cost trade-off as RAM, which forces  applications and the operating system to work within the SRAM (here, $\approx$64--512 KB). Additionally, it is prohibitively expensive to use persistent storage, such as an SD card, for storing temporary results.~\cite{liberis2019neural} Thus, to produce MCU-compatible architectures, we must ensure that the size of activation matrices at its largest remains below the SRAM capacity. 

On an MCU, the network's layers are executed one by one, following a predetermined order, which is a valid topological ordering of layers (nodes) in a computation graph. However, not all topological orders yield the same peak memory usage: whenever the runtime has multiple layers that are ready to be executed (\emph{i.e.} all of their inputs have been computed), the choice of which layer to execute determines which tensor memory buffers are allocated and de-allocated at any given time, which affects the peak memory usage. Thus by considering different topological orders, we can compute the minimum attainable peak memory usage.

Computing the minimum achievable peak memory usage is not straightforward for arbitrary neural network computation graphs. In fact, the few previous works~\cite{banbury2020micronets, fedorov2019sparse} which have explored this problem, resorted to an imprecise under-approximation, which computes the maximum total size of input and output buffers for each operator individually, which does not consider their position within the execution order. This under-approximation is incorrect for networks that contain branches, which are commonplace in state-of-the-art convolutional networks~\cite{sandler2018mobilenetv2, tan2019efficientnet, cheng2019swiftnet, he2016deep}.

Instead, we adopt an accurate peak memory usage metric based on dynamic programming, which efficiently enumerates all topological orderings of the computation graph (feasible for most CNNs) to find the one with the smallest peak memory usage~\cite{liberis2019neural}. In practice, to obtain gradients of layer sizes with respect to peak memory usage, it is sufficient to only run this algorithm on the forward pass of computing the objective and record which tensors formed the memory bottleneck---the peak memory usage is given by the sum of bottleneck tensor sizes, which is differentiable.

In Section~\ref{sec:evaluation}, we will empirically show that the two options for calculating peak memory usage---\emph{precise} and \emph{imprecise}---diverge during pruning on MobileNet-v2 
and ResNet networks, with the imprecise objective causing a violation of resource budget constraints.

\textbf{Model size.} All constant data, including program code and the weights of a neural network, are stored in the Flash memory. The small amount of storage available ($<$ 64--128 KB) imposes a significant constraint on model design.

Our methodology uses quantisation to reduce the storage requirement of a network. Due to its widespread support, we employ affine quantisation~\cite{jacob2018quantization}, which reduces the per-parameter memory requirement to 8 bits; however, other types of quantisation can be used instead. Therefore, the model size objective is the sum of all weight tensor sizes at 1 byte per parameter.

Here, quantisation plays a dual role: in addition to reducing storage usage, it enables execution on the most underpowered microcontrollers, which do not support floating-point computation. Using  -bits, in particular, is a convenient choice for most compute platforms, as it is a natively supported data type: compared to $<$ 8-bit or variable encoding schemes, it does not require any decoding at runtime. 

\subsection{Pruning objectives}
\label{sec:pruning-objectives}

To have fine-grained control over the resource usage of a network, the \emph{width} of each prunable layer, \emph{i.e.} the number of feature maps, channels or units in convolutional and fully-connected layers, is set individually. The remainder of the section will refer to feature maps as \emph{channels}, borrowing CNN terminology. As layers have differing numbers of channels, the algorithm maintains layerwise width multipliers $\vec{\pi} = (\pi_1, \ldots, \pi_L, \ldots)$, where each $\pi_L \in (0; 1]$ determines the fraction of channels to be kept alive (or, conversely, pruned away) within each layer. Each $\pi_L$ is treated as a continuous variable, and the number of channels is determined by rounding down the product of $\pi_L$ and the original layer width.

The optimisation requires a suitable loss $\mathcal{P}$, such that $\frac{\partial \mathcal{P}}{\partial \pi_L}$ can be computed and used to adjust each $\pi_L$ with gradient descent. A linear combination of (a) a resource constraint loss $\mathcal{P}_\textsc{RES}$, which calculates the extent to which the resource budget is violated, and (b) a task loss $\mathcal{P}_\textsc{TSK}$, which encourages maintaining a low classification loss, is used:
\begin{equation}\label{eqn:ch5:pruning-loss}
    \mathcal{P} = \alpha_\textsc{RES} \mathcal{P}_\textsc{RES} +  \alpha_\textsc{TSK} \mathcal{P}_\textsc{TSK}
\end{equation}

Resource budget constraints are, therefore, treated as additional objectives and not as optimisation equality constraints. $\mathcal{P}$ is minimised in parallel to network training as an outer-level optimisation. Therefore, $\mathcal{P}$ is \emph{not} a part of the network training loss and is \emph{not a regulariser} directly. The trade-off hyperparameters control the contribution of the two losses. In practice, the optimisation has low sensitivity to the values of $\alpha$ and the pruning learning rate (established experimentally in Appendix~\ref{apx:hyperparam-sensitivity}).

Note that $\mathcal{P}$ is \emph{not} a part of a network's training loss and is \emph{not a regulariser} directly, since pruning is carried out in parallel to network training as an outer-level optimisation. The hyperparameter $\alpha$ is set depending on the extent of constraint violation and is constant throughout training: we measure the initial values both losses and set $\alpha = r {\mathcal{P}_\textsc{RES}^\textit{init}}/{\mathcal{P}_\textsc{TSK}^\textit{init}}$, where $r$ is usually between $\sfrac{1}{2}$ and $\sfrac{3}{4}$, typically $\sfrac{2}{3}$.

\textbf{Definition of $\mathcal{P}_\text{RES}$.} Differentiable pruning uses the three resource usage objectives: model size, latency (MACs) and peak memory usage. The latter is implemented by the minimal accurate peak memory usage objective that considers layer execution order. 
The objectives can be expressed in terms of, and (piece-wise) differentiated with respect to, the layerwise multipliers $\vec{\pi}$. All objectives are collapsed into one through the use of random scalarisations~\cite{paria2020flexible} to randomly prioritise different objectives at each update iteration ($t$) while maintaining a bias towards constraints that are violated to a greater extent. Coefficients $\vec{\lambda}^t = (\lambda^t_1, \lambda^t_2, \lambda^t_3)$ are sampled randomly $\lambda_i \sim 1 / \textsc{Uniform}(0, 1)$.
\begin{equation}\label{eqn:ch5:resource-loss}
  \mathcal{P}^t_\textsc{RES}(\vec{\pi}) = \max 
  \begin{cases} 
    \lambda^t_1 \ \textsc{PeakMemUsage}(\vec{\pi}) / b_1, \\
    \lambda^t_2 \ \textsc{ModelSize}(\vec{\pi}) / b_2,  \\
    \lambda^t_3 \ \textsc{MACs}(\vec{\pi}) / b_3
  \end{cases}
  - 1
\end{equation}

The optimisation requires the user to specify the desired limit (upper bound) for each resource metric, $b_i$. If randomness is undesirable, the budget-scaled objectives may also be combined via an unweighted sum. For implementation purposes, coefficients $\vec{\lambda}$ are only used to choose the objective within ``max'' and do not scale the gradients. $\mathcal{P}_\textsc{RES}$ is clipped to be $> 0$ after the constraints are satisfied to prevent further compression.

\subsection{Pruning algorithm}
\label{sec:pruning-algorithm}

The pruning of neural network weights is implemented by applying channel-wise binary masks $\mathbf{M}$ to the output of each layer. This effectively sets specific output channels of a layer all to $\mathbf{0}$, allowing the parameters responsible for computing and consuming the masked-out channels to be safely removed from the network. An unpruned neural network is denoted as $f$ and its pruned counterpart as $f^\mathbf{M}$. 

\textbf{Definition of $\mathcal{P}_\text{TSK}$.} Pruning seeks to minimise (or improve upon) the difference between the classification loss of the unpruned and the pruned network on unseen data while only adjusting elements of the binary mask $\mathbf{M}$:
\begin{multline}
\text{argmin}_\mathbf{M} [\mathcal{L}_{ce}(f^\mathbf{M}, \mathcal{D}) - \mathcal{L}_{ce}(f, \mathcal{D})] = \\ \text{argmin}_\mathbf{M} \mathcal{L}_{ce}(f^\mathbf{M}, \mathcal{D})
\end{multline}

where $\mathcal{L}_{ce}$ stands for classification loss (cross-entropy) and $\mathcal{D}$ for a validation dataset. Therefore, the classification loss can be used as task-specific feedback for pruning: 
\begin{equation}
    \mathcal{P}_\textsc{TSK} \stackrel{\text{def}}{=} \mathcal{L}_{ce}(f^\mathbf{M}, \mathcal{D})
\end{equation}

\textbf{Channel importance.} Pruning algorithms typically use a parameter importance (\emph{salience}) criterion to determine which parameters to keep. Both for sparse (unstructured) and structured pruning, popular choices include magnitude-based~\cite{han2015learning}, norm-based ($L_1$ or $L_2$)~\cite{lin2020dynamic}, Hessian-based~\cite{theis2018faster}, batch normalisation (BN)-specific~\cite{liu2017learning}, or gradient flow-based~\cite{lee2018snip, wang2020picking} salience metrics. 

The methodology focuses on convolutional neural networks (CNNs),  assembled out of ``Conv-BatchNorm (BN)-ReLU'' layer sequences, fully connected (FC) layers, and other parameter-free layers. In the former case, the salience $\vec{s}$ is chosen to be scaling coefficients $\vec{\gamma}$ within batch normalisation, and in the latter case, the $L_2$ norm of weights of a neuron. More specifically, the importance of a particular channel/neuron $i$ is $s_i$, given by:
\begin{align*}
\text{Conv-BN-ReLU}(\mathbf{x}) & = \text{ReLU}[ \vec{\gamma}\:\text{Norm}(\mathbf{K} * \mathbf{x}) + \vec{\beta}] \\
s_i & = |\gamma_i| \\
\text{FC}(\mathbf{x}) & = \mathbf{Wx} + \vec{b} \\
s_i & = ||\mathbf{W}_{:,i}||_2
\end{align*}
where $\mathbf{x}$ is the input tensor, $\mathbf{K}$, $\mathbf{W}$ and $\vec{b}$ are parameter tensors, $*$ is a convolution operation and ``Norm'' is a mean-zero variance-one normalisation function, and $\vec{\gamma}$ and $\vec{\beta}$ are learned batch normalisation scaling and offset parameters.

\textbf{Pruning by salience.} Suppose that within a layer $L$ with $C$ output channels, a proportion $\pi_L$ is set to be kept and not pruned. During pruning, all channels are ranked according to their salience ($\vec{s_L}$) and top $\pi_L$ ($\pi_L^{th}$ quantile) are kept (entries in $\textbf{M}$ are set to 1) and bottom $1 - \pi_L$ are discarded ($\textbf{M}$ set to 0).

Therefore, for a given layer $L$, $\textbf{M}$ is computed by \emph{a hard threshold} function, denoted $M$, of channel salience, with the salience threshold $\tau_L$ picked to satisfy the constraint that only $\pi_L$ proportion of channels have salience $s_{L,i} > \tau_L$. Remaining text refers to $\vec{s_L}$ and $s_{L, i}$ as $\vec{s}$ and $s_i$, respectively, making the layer implicit for brevity.

\textbf{Differentiable pruning.} Gradient update to $\pi_L$ is calculated assuming a continuous relaxation of the hard threshold function on the backwards pass only. Following DSA~\cite{ning2020dsa}, we use a sigmoid function in the log-domain of saliences, which is differentiable with respect to $\tau_L$: 
\begin{equation}\label{eqn:smooth-mask}
    M(s_i, s_0) = \frac{1}{1 + e^{-[\ln s_i - \ln s_0]}} = \frac{1}{1 + s_0 / s_i}
\end{equation}
and enforce a constraint that the proportion of kept channels should equal the desired channel width multiplier $\pi_L$:
\begin{equation} \label{eqn:pi-constraint}
  \frac{1}{C} \sum_j^C M(s_j, s_0) = \pi_i
\end{equation}


Differentiating Eqn.~\ref{eqn:pi-constraint} w.r.t. $\pi_i$ and rearranging gives:
\begin{align}
    \frac{1}{C} \sum_j^C \frac{\partial M(s_j, s_0)}{\partial s_0} \frac{\partial s_0}{\partial \pi_i} = 1 \\
    \frac{\partial s_0}{\partial \pi_i} = \frac{C}{\sum_j^C \frac{\partial M(s_j, s_0)}{\partial s_0}} \label{eqn:pi-constraint-deriv}
\end{align}

This is used to connect task loss gradients with respect to the mask $M$ back to each $\pi_L$:
\begin{multline}\label{eqn:deriv-convert}
    \frac{\partial \mathcal{P}_\textsc{TSK}}{\partial \pi_i} = [\sum_j^C \frac{\partial \mathcal{P}_\textsc{TSK}}{\partial M(s_j, s_0)}\frac{\partial M(s_j, s_0)}{\partial s_0}] \frac{\partial s_0}{\partial \pi_i} \\
    \stackrel{{\text{Eqn. \ref{eqn:pi-constraint-deriv}}}}{=} C \sum_j^C \frac{\partial \mathcal{P}_\textsc{TSK}}{\partial M(s_j, s_0)}\frac{\partial M(s_j, s_0)/\partial s_0}{\sum_k^C \partial M(s_k, s_0)/\partial s_0}
\end{multline}

\textbf{Focus regime for bottleneck objectives.} One out of three resource usage objectives considered is a \emph{bottleneck objective}: only a few layers (the bottleneck) contribute to peak memory usage at a time. This differs from model size and MACs, to which all layers contribute to a certain extent. 

As per Equation~\ref{eqn:ch5:pruning-loss}, gradients from both $\mathcal{P}_\textsc{TSK}$ and $\mathcal{P}_\textsc{RES}$ are combined to update $\vec{\pi}$, which may needlessly shrink layers that are not in the bottleneck when other resource constraints are already satisfied. To prevent this, $\pi_L$s is updated only for bottleneck layers, \emph{i.e.} layers for which $\frac{\partial \mathcal{P}_\textsc{RES}}{\partial \pi_L} > 0$. 

\begin{figure}[h]
    \centering
    \includegraphics[scale=0.9]{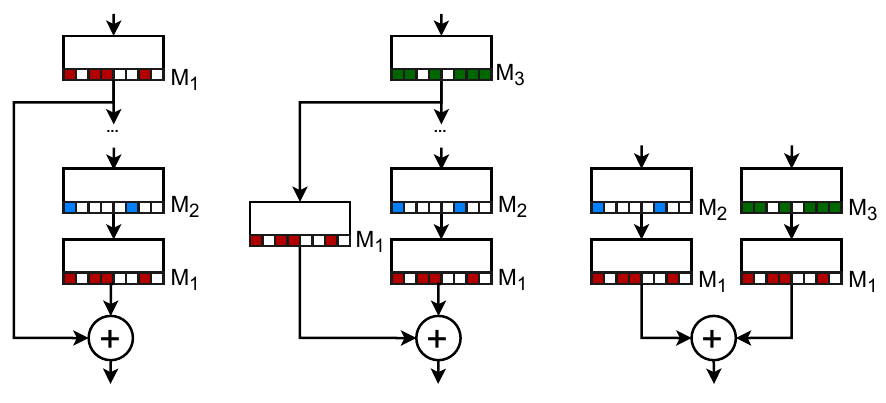}%
    \caption{Three examples of pruning over \texttt{Add}, such as residual connections. All summands share the same pruning mask.}
    \label{fig:res-pru}
\end{figure}

\textbf{Pruning residual connections.} 
Many modern CNNs feature multiple processing paths aggregated using addition. To correctly preserve the addition of pruned feature maps, their surviving (unpruned) channels must have matching positions, \emph{i.e.} their pruning masks have to be equal~\cite{ning2020dsa,van2020single}. It is also possible to work around this requirement with padding and channel rearrangement, albeit at extra implementation overhead~\cite{lemaire2019structured}.

Figure~\ref{fig:res-pru} shows examples of tying pruning masks in practice, including residual connections~\cite{he2016deep}. Groups of layers that share the pruning mask are pre-computed in advance using connected components search; the combined salience is the maximum of saliences of each layer. 

\subsection{Implementation}

Pruning iterations are interleaved with model training steps, providing compression with no subsequent fine-tuning: the accuracy loss from the marginal reduction of layer sizes can be recovered before the next pruning iteration.

\begin{algorithm}[h]
\caption{A high-level implementation of the update procedure for width multipliers $\pi$ (\textsc{WidthUpdate}). $f$, $\mathcal{P}_\textsc{RES}$, $\mathcal{P}_\textsc{TSK}$, $\mathcal{D}$, $\vec{s}$, $\tau_L$ as defined in Sections~\ref{sec:pruning-objectives},~\ref{sec:pruning-algorithm}. \textsc{Diff}($f$, $x$) computes $\frac{\partial f}{\partial x}$. \textsc{Salience} and \textsc{Percentile} are self-descriptive helper functions.}
\label{alg:pruning-algo}
\begin{algorithmic}
\vspace{1mm}
\Function{WidthUpdate}{$f$, $\vec{\pi}$, $\mathcal{D}$}
\State $\mathbf{M}, \frac{\partial \mathbf{M}}{\partial \tau_L} \gets$ [\Call{Mask}{Layer$_i$, $\pi_L$} $\textbf{for } \text{Layer}_i \in f$]
\State $\frac{\partial \mathcal{P}_\textsc{TSK}}{\partial \mathbf{M}} \gets$ \Call{Diff}{$\mathcal{P}_{ce}(f^\mathbf{M}, \mathcal{D})$, $\mathbf{M}$}

\State $\frac{\partial \mathcal{P}_\textsc{TSK}}{\partial \vec{\pi}} \gets$ use Equation~\ref{eqn:deriv-convert} with $\frac{\partial \mathcal{P}_\textsc{TSK}}{\partial \mathbf{M}}$ and $\frac{\partial M}{\partial \tau_L}$ 

\State $\frac{\partial \mathcal{P}_\textsc{RES}}{\partial \vec{\pi}} \gets$ \Call{Diff}{$\mathcal{P}_\textsc{RES}(\vec{\pi})$, $\vec{\pi}$} \Comment{Equation~\ref{eqn:ch5:resource-loss}}
\State $\frac{\partial \mathcal{P}}{\partial \vec{\pi}} \gets \alpha_\textsc{RES} \frac{\partial \mathcal{P}_\textsc{RES}}{\partial \vec{\pi}} + \alpha_\textsc{TSK} \frac{\partial \mathcal{P}_\textsc{TSK}}{\partial \vec{\pi}}$ \Comment {Equation~\ref{eqn:ch5:pruning-loss}}

\State $\vec{\pi} \gets \vec{\pi} - \eta_\pi \frac{\partial \mathcal{P}}{\partial \vec{\pi}}$ \Comment{SGD update with learn. rate $\eta_\pi$}
\EndFunction
\vspace{1mm}
\Function{Mask}{L, $\pi_L$}
\State $\vec{s} \gets $\Call{Salience}{L} \Comment{as per Section~\ref{sec:pruning-algorithm}}
\State $\tau_L \gets $\Call{Percentile}{$\vec{s}$, $\pi_L$} \Comment{$\pi_L^{\text{th}}$ percentile of $\vec{s}$}
\State $\mathbf{M} \gets (\vec{s} > \tau_L)$
\State $\frac{\partial M}{\partial \tau_L} \gets$ \Call{Diff}{$M(\vec{s}, \tau_L)$, $\tau_L$} \Comment{Equation~\ref{eqn:smooth-mask}}
\State \Return $\mathbf{M}, \frac{\partial M}{\partial \tau_L}$
\EndFunction
\end{algorithmic}
\end{algorithm}

Layer masks and layerwise multipliers $\vec{\pi}$ are updated using functions in Algorithm~1 every 20 training steps: $\vec{\pi}$ is updated using stochastic gradient descent (SGD) with learning rate $\eta_\pi$ within \textsc{WidthUpdate}, followed by recalculating masks using \textsc{Mask} (without computing $\frac{\partial M}{\partial \tau_L}$). To enforce $\pi_L \in (0, 1]$, $\vec{\pi}$ is stored as an unconstrained real variable vector internally with a scaled ``exp'' function applied to it ($\pi_L$ starts at 1.0 and, by design, can only decrease in value). The flat shape of ``exp'' for small values of the underlying variable also discourages shrinkage of small $\pi_i$'s, akin to $L_2$ regularisation.

Pruning stops---that is, channel width multipliers $\vec{\pi}$ are no longer updated---when the user-specified resource usage bounds have been reached. However, channel masks $\mathbf{M}$ are not frozen until the final quantisation-aware training stages.

\section{Evaluation}
\label{sec:evaluation}

Differentiable pruning for microcontrollers is empirically evaluated as follows:
\begin{enumerate}
    \item The utility of resource usage objectives within pruning is tested through an ablation study to show that this feedback is essential for targeting microcontroller-specific bottlenecks.
    \item An accurate peak memory usage objective is compared to memory usage metrics used in prior work (if any) to show that the precise calculation is required for correctly allocating the SRAM budget.
    \item Pruning is compared to the uniform scaling of layers on MobileNet-v2~\cite{sandler2018mobilenetv2} to show that pruning allocates resources more efficiently.
    \item Pruned models for fourteen architecture and task pairs are presented, with an improved classification accuracy, resource usage or time taken for compression, compared to related network pruning and microcontroller-specialist work.
    \item The low sensitivity of the methodology to feedback trade-off hyperparameters $\alpha$ and pruning learning rate $\eta_\pi$ is established (Appendix~\ref{apx:hyperparam-sensitivity}).
\end{enumerate}

Further analysis and discussion of the applicability of this methodology to real-world TinyML tasks are presented in Section~\ref{sec:discussion}.

\textbf{Baseline comparisons.} 
Differentiable pruning is tested on Speech Commands~\cite{warden2018speech}, CIFAR-10~\cite{krizhevsky09learning}, Visual Wake Words (VWW)~\cite{chowdhery2019visual} (``person'' \emph{vs} object classification)\footnote{Full 3-channel colour inputs are used (does not violate the SRAM constraint, unlike in \citet{banbury2020micronets}); full ``val'' split is used for final evaluation.} and ImageNet (ILSVRC'12) datasets. The pruning is applied to a wide range of architectures: VGG-16~\cite{simonyan2014very}, ResNet-18~\cite{he2016deep}, MobileNet-v2~\cite{sandler2018mobilenetv2}, EfficientNet-B0~\cite{tan2019efficientnet}, DS-CNN~\cite{zhang2017hello}, RES-8/15 (ResNet-like)~\cite{tang2018deep}, MCUNet\footnote{For MCUNet, the architectures instantiated at full channel conv. counts (``width''), excl. MCUNet-L (already at full width), and are pruned to the resource usage of models presented by the authors.}~\cite{lin2020mcunet} and EtinyNet~\cite{xu2022etinynet}\footnote{The architectures use downsampled input, instead of downsampling in the first pooling layer, to minimise PMU.}.

To the best of our knowledge, this work is the first network pruning methodology constructed to optimise for microcontroller resource bottlenecks. As there are no directly comparable microcontroller-specialist pruning methods, the comparisons are performed against, firstly, prior differentiable larger-scale network pruning methods and, secondly, microcontroller NAS systems for TinyML tasks. More specifically, the evaluation aims to: 
\begin{itemize}
    \item[(a)] Verify that the compressed models have comparable or better resource usage than GPU-/mobile-scale differentiable budgeted pruning methods, DSA~\cite{ning2020dsa} and AutoCompress~\cite{liu2019autocompress} (auto filter pruning), for CIFAR-10.
    \item[(b)] Show that this methodology can produce comparable or better models than the closest prior MCU specialist work, MicroNets~\cite{banbury2020micronets} and MCUNet~\cite{lin2020mcunet}, for other datasets, and do so faster due to the use of pruning.
    \item[(c)] Verify that differentiable pruning would not deliver high-performant models without being guided by the microcontroller-specific resource usage objectives.
\end{itemize}

Unless otherwise specified, the baseline models have been reimplemented, and their resource usage was recalculated within the same framework\footnote{Standard execution strategy, like in the TensorFlow Lite Micro runtime, with support for accumulating into one of the summands of the `Add' operator; no other non-standard system-level improvements (4-bit quantisation, custom kernel implementations, etc.) are considered.} to facilitate a fair comparison. All models use the same training and data augmentation protocols and are quantised to 8-bit precision. Distillation, fine-tuning using the validation set, or other adjustments which may confound the results are not used. For ``time'' comparison with NAS methods, the computation required to train a supernetwork model and subsequently fine-tune the discovered smaller model is compared against the computation needed for a single training run of the full model (this work), with the amounts of computation estimated using MAC operation counts. 

\subsection{Ablation studies}

\textbf{Task-only or resource-only pruning.} To explore the impact of the two feedback losses, pruning can be performed using only the task loss\footnote{Ordinarily, the task-specific loss only provides ``negative'' feedback against the reduction of $\pi_L$ for salient layers. For this experiment, however, we allow only positive $\sfrac{\partial \mathcal{P}_\textsc{TSK}}{\partial \pi_L}$, as there is no resource usage loss feedback to reduce $\pi_L$.} ($\mathcal{P}_\textsc{RES} = 0$) or only the resource-constraint loss ($\mathcal{P}_\textsc{TSK} = 0$). Table \ref {tab:loss-ablation} compares two MobileNet-v2~\cite{sandler2018mobilenetv2} networks on the Visual Wake Words (VWW) dataset, pruned using either configuration. 

\begin{table}[h]
    \small
    \centering
    \begin{tabular}{lr}
         \toprule
         \textbf{MobileNet-v2 configuration for VWW} & \textbf{Accuracy} \\
         \midrule
         res. 50$\times$50, only $\mathcal{P}_\textsc{TSK}$ & 70.65\% \\ 
         res. 50$\times$50, only $\mathcal{P}_\textsc{RES}$ & 83.11\% \\
         res. 50$\times$50, both (58.0\% sparsity) & \textbf{83.80\%} \\
         \midrule
         res. 160$\times$160, only $\mathcal{P}_\textsc{TSK}$ & 87.90\% \\
         res. 160$\times$160, only $\mathcal{P}_\textsc{RES}$ & 88.80\% \\
         res. 160$\times$160, both (36.4\% sparsity) & \textbf{89.05\%} \\
         \bottomrule
    \end{tabular}
    \caption{Loss ablation study results for the MobileNet-v2 model trained on Visual Wake Words dataset (50$\times$50 and 160$\times$160 input resolution). The target is the resource usage of the MicroNets VWW-1 and VWW-2 baselines; sparsity refers to the proportion of removed channels (regardless of their relative size). \textbf{Both losses are required to achieve top accuracy.}}
    \label{tab:loss-ablation}
\end{table}

The data shows that both losses are required to correctly prioritise unimportant and resource-expensive channels for removal, although tested models can largely successfully recover from resource-driven pruning alone:

\begin{itemize}
\item Pruning with $\mathcal{P}_\textsc{TSK}$ relegates this work to a class of resource-unaware structured pruning methods. More channels are expected to be pruned away until the desired resource usage is eventually reached due to a reduced ability to target computationally-expensive layers first, resulting in lower accuracy overall.

\item Pruning with $\mathcal{P}_\textsc{RES}$ results in an iterative layer adjustment in proportion to the constraint violation, with no awareness of which layers contribute more to generalisation. The lack of protection of more salient layers (no $\mathcal{P}_\textsc{TSK}$) results in lower accuracy overall, although the accuracy differences are minor, suggesting a greater ability to compensate for removed channels when sparsity is low. In general, one cannot predict the resulting level of sparsity compared to using both losses, as it depends on the relative contribution of each channel to resource usage.
\end{itemize}

\textit{Summary.} Adding microcontroller-specific resource usage feedback to pruning, one of the main contributions of this work, is justified: parameter importance-based pruning alone is inadequate for obtaining accurate highly-compressed neural networks.

\newcommand{\suptag}[1]{\tiny\raisebox{3pt}{ #1}}
\newcommand{\CC}[1]{\cellcolor{lightergray}{#1}}
\begin{table*}
\caption{MCU-sized architectures discovered by differentiable network pruning \emph{vs} others. The last column compares a baseline to our pruned model in the group (in light grey). ``Time'' is the relative total training time (search or pruning), estimated using MACs as a proxy for training time. `INT8' represents a reimplemented quantised model, `FP32' denotes a full precision model as reported by its authors (``\textit{unk.}'' is unreported data). Differentiable pruning can discover MCU-sized models on par or better than the related work with negligible overhead (or faster).}
\label{tab:diffpru-results}
\centering
\small
\begin{tabular}{llrrrrrr}
\toprule
\textbf{Backbone} & \textbf{Model} & \textbf{Time} & \textbf{Acc. (\%)} & \textbf{Size} & \textbf{PMU} & \textbf{MACs} & \textbf{Difference} \\
\midrule

\multirow{2}{*}{\makecell[l]{VWW (50x50) on \\ MobileNet v2}} 
& MobileNet v2 (unpruned)\suptag{INT8} & $\times$1.00 & 84.93 & 2.32 M & 76.2 KB & 22.58 M & $\uparrow$ MAC 8.9$\times$ \\
& MicroNets VWW-2\suptag{INT8} & $\times$1.15 &  83.50 & 103 K & 27.9 KB & 3.383 M & $\uparrow$ MAC 1.3$\times$ \\
\rowcolor{lightergray} \cellcolor{white} & \textit{ours \#1 (matching VWW-2)}\suptag{INT8} & $\times${1.00} & 83.40 & {101 K} & {25 KB} & \textbf{2.528 M} &  \\
\rowcolor{lightergray} \cellcolor{white} & \textit{ours \#2 (matching VWW-2)}\suptag{INT8} & $\times${1.00} & 83.80 & {101 K} & {27.8 KB} & \textbf{3.342 M} &  \\
\cmidrule{2-8} 
\multirow{2}{*}{\makecell[l]{VWW (50x50) on \\ EfficientNet-B0}} 
& EfficientNet-B0 (unpruned)\suptag{INT8}\kern-1em & $\times$1.00 & 86.23 & 366 K & 300 KB & 9.468 M & $\uparrow$ PMU 11$\times$\\
& \CC{\textit{ours (m/ VWW-2, early term.)}\suptag{INT8}} & \CC{\textbf{$\times$0.49}} & \CC{83.34} & \CC{95.0 K} & \CC{\textbf{27.8 KB}} & \CC{3.245 M} & \CC{} \\
\midrule
\multirow{2}{*}{\makecell[l]{VWW (160x160) \\ on MobileNet v2}} 
& MobileNet v2 (unpruned)\suptag{INT8} & $\times$1.00 & 90.13 & 2.32 M & 768 KB & 164.8 M & $\uparrow$ MAC 3.2$\times$ \\
& MicroNets VWW-1\suptag{INT8} & $\times$1.43 & 88.49 & 616 K & 200 KB & 71.58 M & $\uparrow$ MAC 1.4$\times$ \\
\rowcolor{lightergray} \cellcolor{white} & \textit{ours (matching VWW-1)\suptag{INT8}} & $\times${1.00} & 89.05 & 606 K & 198 KB & \textbf{58.33 M} & \\
\cmidrule{2-8} 
\multirow{2}{*}{\makecell[l]{VWW (160x160) \\ on EfficientNet-B0}} 
& EfficientNet-B0 (unpruned)\suptag{INT8}\kern-1em & $\times$1.00 & 89.39 & 3.66 M & 768 KB & 187.6 M & $\uparrow$ Size 6$\times$\\
& \CC{\textit{ours (matching VWW-1)}\suptag{INT8}} & \CC{\textbf{$\times$1.00}} & \CC{88.77} & \CC{\textbf{601 K}} & \CC{199 KB} & \CC{52.13 M} & \CC{} \\
\midrule
\multirow{2}{*}{\makecell[l]{CIFAR10 on \\ VGG-16}} 
& VGG-16 (unpruned)\suptag{INT8} & $\times${1.00} & 93.52 & 14.7 M & 131 KB & 313.3 M & $\uparrow$ Size 80$\times$ \\
& A/C (from 93.7\% acc.)\suptag{FP32} & $\times${1.00} & 88.78 & 311 K & \emph{unk.} & 22.38 M & $\downarrow$ Acc. -2.4\% \\
\rowcolor{lightergray} \cellcolor{white} & \textit{ours \#1 (matching A/C)}\suptag{INT8} & $\times${1.00} & \textbf{91.15} & {309 K} & 52.2 KB & 22.38 M & \\
\rowcolor{lightergray} \cellcolor{white} & \textit{ours \#2 (matching A/C)}\suptag{INT8} & $\times${1.00} & 90.38 & \textbf{184 K} & 62.5 KB & 22.15 M & \\
& DSA (from 93.5\% acc.)\suptag{FP32} & $\times${1.00} & 90.16 & \emph{unk.} & \emph{unk.} & 15.35 M & -- \\
\rowcolor{lightergray} \cellcolor{white} & \textit{ours (matching DSA)}\suptag{INT8} & $\times${1.00} & 90.36 & 834 K & 34.8 KB & 15.28 M & \\
\cmidrule{2-8} 
\multirow{2}{*}{\makecell[l]{CIFAR10 on \\ ResNet-18}} 
& DSA (from 94\% acc.)\suptag{FP32} & $\times${1.00} & 93.10 & \emph{unk.} & \emph{unk.} & 97.51 M & -- \\
& \CC{\textit{ours (m/ DSA from 95\% acc.)}\suptag{INT8}} & \CC{$\times${1.00}} & \CC{94.49} & \CC{2.44 M} & \CC{88.0 KB} & \CC{97.20 M} & \CC{} \\
& \CC{\textit{ours (MCU-sized)}\suptag{INT8}} & \CC{$\times${1.00}} & \CC{91.98} & \CC{256 K} & \CC{81.9 KB} & \CC{29.58 M} & \CC{} \\
\midrule
\multirow{2}{*}{\makecell[l]{KWS on \\ DS-CNN}} 
& DS-CNN (MN-L, unpruned)\suptag{INT8}\kern-1em & $\times${1.00} & 96.79 & 582 K & 170 KB & 74.27 M & $\uparrow$ Size 14\% \\
& MicroNets KWS-L\suptag{INT8} & $\times${1.89} & 96.56 & 512 K & 170 KB & 65.75 M & $\uparrow$ Speed 89\% \\
\rowcolor{lightergray} \cellcolor{white} & \textit{ours (matching KWS-L)}\suptag{INT8} & \textbf{$\times${1.00}} & 96.56 & 511 K & 159 KB & 65.47 M &  \\
& DS-CNN (MN-M, unpruned)\suptag{INT8}\kern-1em & $\times${1.00} & 96.36 & 420 K & 170 KB & 54.61 M & $\uparrow$ MAC 3.6$\times$ \\	
& MicroNets KWS-M\suptag{INT8} & $\times${1.29} & 95.73 & 117 K & 86.1 KB & 15.58 M & $\uparrow$ Speed 29\% \\
\rowcolor{lightergray} \cellcolor{white} & \textit{ours (matching KWS-M)}\suptag{INT8} & \textbf{$\times${1.00}} & 96.03 & 115 K & 83.0 KB & 15.56 M & \\
& MicroNets KWS-S\suptag{INT8} & $\times${1.15} & 95.40 & 63.6 K & 51.7 KB & 8.351 M & $\uparrow$ Speed 15\% \\
\rowcolor{lightergray} \cellcolor{white} & \textit{ours (matching KWS-S)}\suptag{INT8} & \textbf{$\times${1.00}} & 95.75 & 61.9 K & 49.8 KB & 8.342 M & \\
\cmidrule{2-8} 
\multirow{2}{*}{\makecell[l]{KWS on \\ RES-8/15}} 
& RES-15 (unpruned)\suptag{INT8} & $\times${1.00} & 96.48 & 240 K &  66.1 KB & 116.4 M & $\uparrow$ MAC 7.8$\times$\\
& \CC{\textit{ours (matching KWS-M)}\suptag{INT8}} & \CC{$\times${1.00}} & \CC{95.42} & \CC{32.4 K} & \CC{27.4 KB} & \CC{\textbf{14.96 M}} & \CC{} \\
\cmidrule{2-8}
& RES-8 (unpruned)\suptag{INT8} & $\times${1.00} & 93.25 & 112 K & 23.7 KB & 4.162 M & $\uparrow$ PMU 73\%\\
\rowcolor{lightergray} \cellcolor{white} & \textit{ours (matching KWS-S)}\suptag{INT8} & $\times${1.00} & 92.78 & 63.2 K & \textbf{13.7 KB} & 2.354 M & \\
\midrule
\multirow{2}{*}{\makecell[l]{ImageNet on \\ MCUNet}} 
& MCUNet-S\suptag{INT8} & $\times${1.25} & 59.78 & 748 K & 333 KB & 67.41 M & $\uparrow$ Speed 25\% \\
& \CC{\textit{ours (matching MCUNet-S)}\suptag{INT8}} & \CC{\textbf{$\times${1.00}}} & \CC{59.78} & \CC{740 K} & \CC{256 KB} & \CC{66.59 M} & \CC{} \\
& MCUNet-M\suptag{INT8} & $\times${1.26} & 60.98 & 756 K & 341 KB & 81.94 M & $\uparrow$ Speed 26\% \\
\rowcolor{lightergray} \cellcolor{white} & \textit{ours (matching MCUNet-M)}\suptag{INT8} & \textbf{$\times${1.00}} & 60.92 & \textbf{647 K} & 280 KB & 81.66 M & \\
\cmidrule{2-8}
\multirow{2}{*}{\makecell[l]{ImageNet on \\ EtinyNet}} 
& EtinyNet $\times 1.0$ (unpruned)\suptag{INT8} & $\times${1.00} & 60.18 & 979 K & 201 KB & 105.9 M & $\uparrow$ MAC 1.7$\times$ \\
& EtinyNet $\times 0.75$\suptag{INT8} & $\times${1.00} & 55.75 & 660 K & 151 KB & 63.35 M & $\downarrow$ Acc. -0.8\% \\
& \CC{\textit{ours (matching $\times 0.75$)}\suptag{INT8}} & \CC{\textbf{$\times${1.00}}} & \CC{56.55} & \CC{638} K & \CC{149 KB} & \CC{63.32 M} & \CC{} \\
\bottomrule
\end{tabular}
\end{table*}

\textbf{Usefulness of peak memory usage.}  One of the essential MCU resource constraints to be addressed, which does not similarly manifest on other platforms, is peak memory usage. Differentiable pruning uses the accurate minimal peak memory usage metric, which optimises layer execution order. Three pruning configurations are considered for testing the contribution of the memory usage objective: (a) no peak memory usage objective, leaving only latency and model size; (b) an imprecise under-approximating peak memory usage objective, used in prior work~\cite{banbury2020micronets,fedorov2019sparse}; (c) the full set of resource objectives. The results are expected to show a gap between the precise and the imprecise measurements; only pruning with the precise objective should produce a model whose true peak memory usage lies within the memory budget.

\begin{figure}[h]
    \centering
    \includegraphics[scale=0.85]{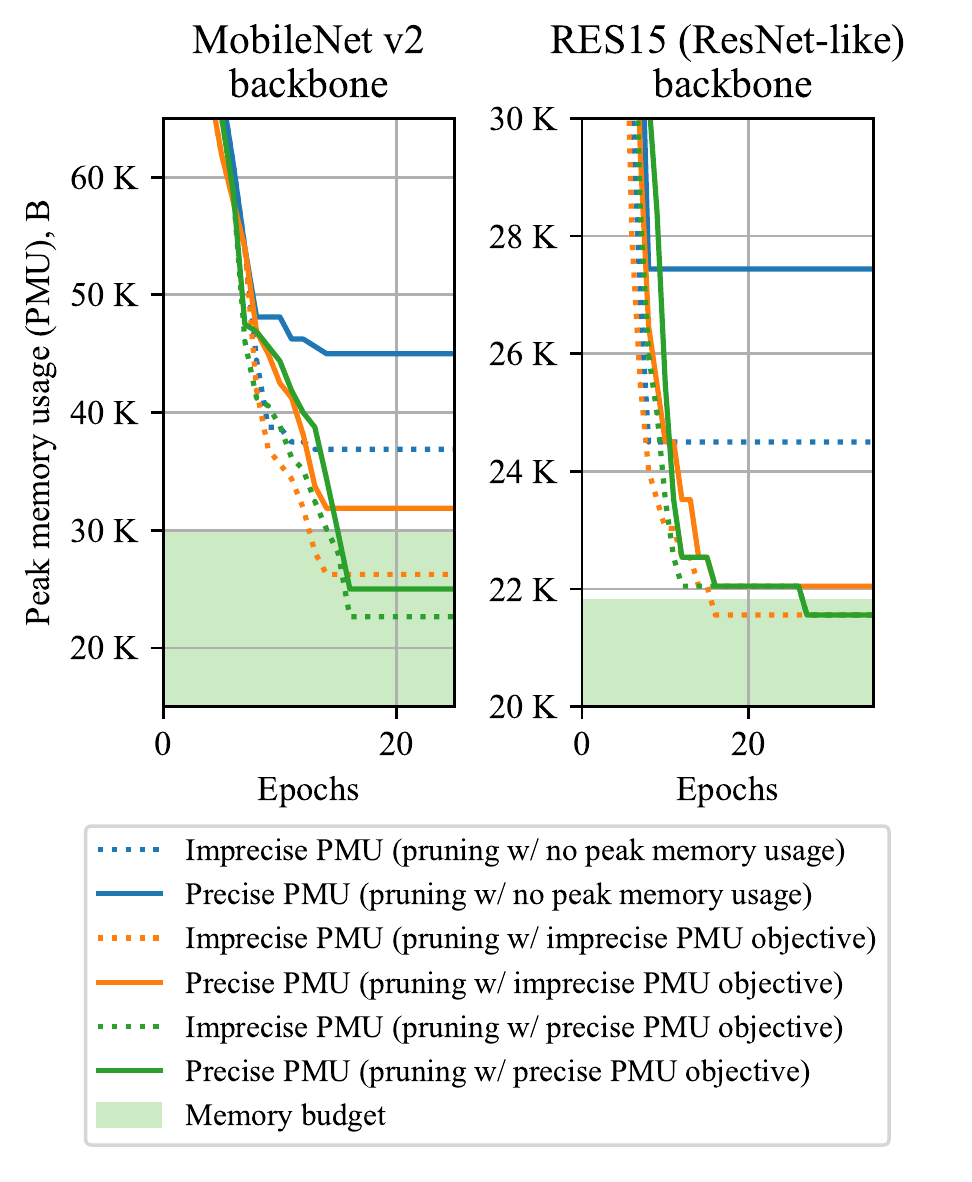}%
    \caption{Precise (solid) and imprecise (dashed) peak memory usage on MobileNet v2 and RES15 (ResNet-like) backbones. The three colours represent different pruning experiments which have different PMU objectives: the imprecise calculation, the precise calculation and no PMU objective. In all cases, the imprecise objective underapproximates the true peak memory usage.}
    \label{fig:pmu-objectives}
\end{figure}

Figure~\ref{fig:pmu-objectives} shows the evolution of peak memory usage (PMU) during pruning. The data shows that: (a) using an under-approximation to PMU would cause pruning to be terminated early, potentially while the true memory usage still exceeds the memory budget; (b) the two PMU measures may diverge as the network architecture changes; (c) using PMU as a pruning objective is necessary to ensure that the final memory usage lays within the budget; that is, the peak memory usage requirement is \emph{not} satisfied via other objectives.

\textit{Summary.} Targeting model size or latency (MACs) alone, as done in some prior GPU- or mobile-level resource-aware pruning work~\cite{gordon2018morphnet,ning2020dsa}, is insufficient for obtaining microcontroller-compatible models, as peak memory usage would not be sufficiently reduced. Using an accurate memory usage objective is essential for both correctly identifying which tensors contribute to the memory bottleneck and determining when the desired resource usage has been reached to stop pruning.

\textbf{Comparison to uniform scaling.} Many popular CNNs, such as MobileNet-v2 and EfficientNet, are offered at a range of manually-engineered resource usage (or size) \emph{vs} accuracy trade-off points to support a variety of resource footprints. These are typically defined by uniformly scaling the number of convolutional channels across the network layers. Similarly, pruning achieves a range of trade-off points and is expected to perform better due to the ability to adjust each layer individually.

To compare pruning and uniform scaling, different scales of the MobileNet-v2 model are considered---$\times$0.75, $\times$0.50, $\times$0.25---as defined by the authors.
Pruning is applied to the base architecture ($\times$1.0) to produce a network that matches or improves upon the resource usage of each of the uniformly scaled models while preserving or improving upon the classification accuracy. 

\begin{figure}[h]
    \centering
    \includegraphics[scale=0.8]{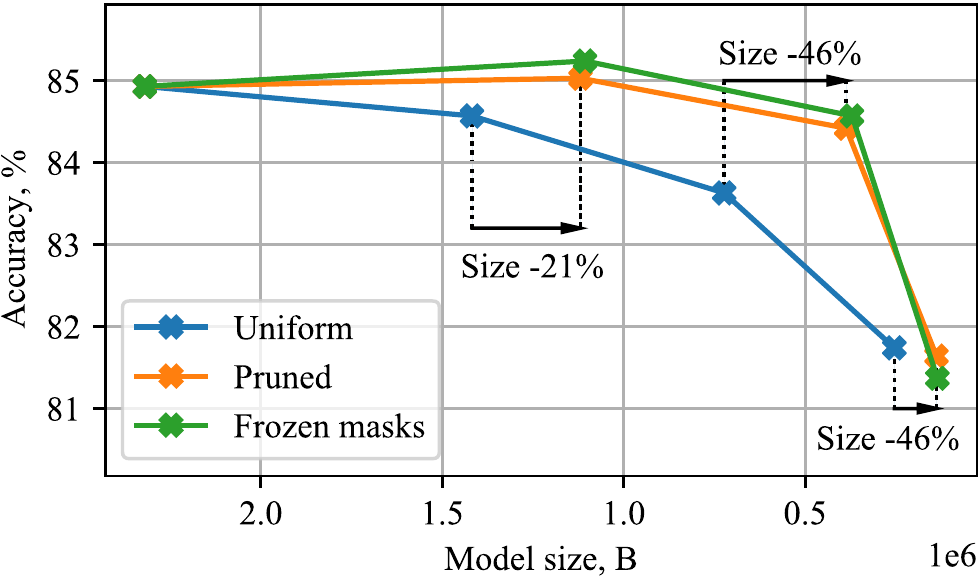}%
    \caption{The Pareto front of MobileNet-v2 models applied to the VWW (50x50) dataset. The models are compressed using uniform scaling, pruning, and pruning with masks frozen as soon as layer sizes have been learned (for training efficiency). Early termination refers to ``freezing'' pruning masks upon convergence. Pruning yields an improved Pareto front of models.}
    \label{fig:vww-mbnet-pru}
\end{figure}

Figure~\ref{fig:vww-mbnet-pru} shows the accuracy and resource usage Pareto front for scaled or pruned MobileNet-v2 models. The results confirm the hypothesis that pruning can find a more efficient resource budget allocation, reducing the size of the final model by up to 46\%. Pruning is known to have a regularising effect at low pruning ratios~\cite{hoefler2021sparsity}, resulting in a boost in performance compared to the baseline model at $\times$0.75 scaling.

\textit{Summary.} Computational resource budget can be allocated more efficiently by pruning instead of uniformly scaling network layers. Differentiable pruning improves the classification accuracy \emph{vs} resource usage Pareto front.

\subsection{Discovered models}
\label{sec:discovered-models}

Table~\ref{tab:diffpru-results} presents several low-footprint models obtained by differentiable pruning and using related work, as per evaluation settings described earlier. The methodology targets more capable hardware than previously considered in μNAS: ``mid-to-high-tier'' MCUs with $>$ 64 kB of storage and $>$ 32 kB of memory. For each dataset and architecture (backbone) combination, we consider the relative amount of computation required to obtain the model (``compression time''), the classification accuracy, and resource usage of (a) a full unpruned model (where illustrative), (b) a baseline compressed model from prior work and (c) a pruned model obtained using this methodology. `INT8' represents a reimplemented quantised model, `FP32' denotes a full precision model as reported by its authors; ``\textit{unk.}'' is unreported data.

The data shows that differentiable pruning improves upon the accuracy, resource usage or relative speed of compression compared to baseline methods. Most notably, this work (a) improves key resource usage of neural networks up to 80$\times$; (b) has negligible overhead compared to prior microcontroller-specific NAS methodologies, and can even reduce the amount of computation needed to train the network with early termination; (c) produces compressed models with matching or improved resource usage by up to 40\% compared to prior work. 

\textit{Summary.} Differentiable pruning produces superior models for a range of microcontroller resource usage budgets compared to related work without any task-specific adjustment. 

\section{Discussion}
\label{sec:discussion}

The following section discusses points relating to the practical application of differentiable pruning:
\begin{itemize}
    \item Terminating pruning at the earliest possible opportunity can reduce the amount of computation required for training, compared to training without pruning.
    \item Two pruned models are analysed using power use and latency measurements on two MCU development boards to establish the viability of on-device inference.
    \item The two models are analysed layer-wise to identify which layers are targeted by pruning to satisfy the resource budget (Appendix~\ref{apx:arch-anal}).
    \item Differentiable pruning is broadly applicable to tasks typically tackled by the ``smart'' software running on embedded, wearable and IoT devices.
\end{itemize}

\subsection{Early termination}
\label{sec:ch5:discussion-early-term}

The stages of network training at which the layer sizes $\vec{\pi}$ are learned can be controlled by adjusting the learning rate and the epoch at which the learning starts. Once the architecture satisfies the resource constraints, the learning of $\vec{\pi}$ stops; however, the pruning masks $\textbf{M}$ can continue to change. Previously pruned channels can be re-enabled if the pruning threshold $\tau_L$ falls below salience $s_i$. This continuing change is akin to spatial Dropout regularisation~\cite{tompson2014efficient}, which may either improve or hinder performance.

The computation required for training can be reduced by switching to training the pruned model once $\vec{\pi}$ has converged (or soon after). This computation saving can be maximised if (a) pruning starts early in the training process and (b) converges quickly (high learning rate $\eta_\pi$), and (c) the regularising effect of continuing mask change can be forgone without significant impact to the final performance of the model.

\begin{figure}[h]
    \centering
    \includegraphics[scale=0.75]{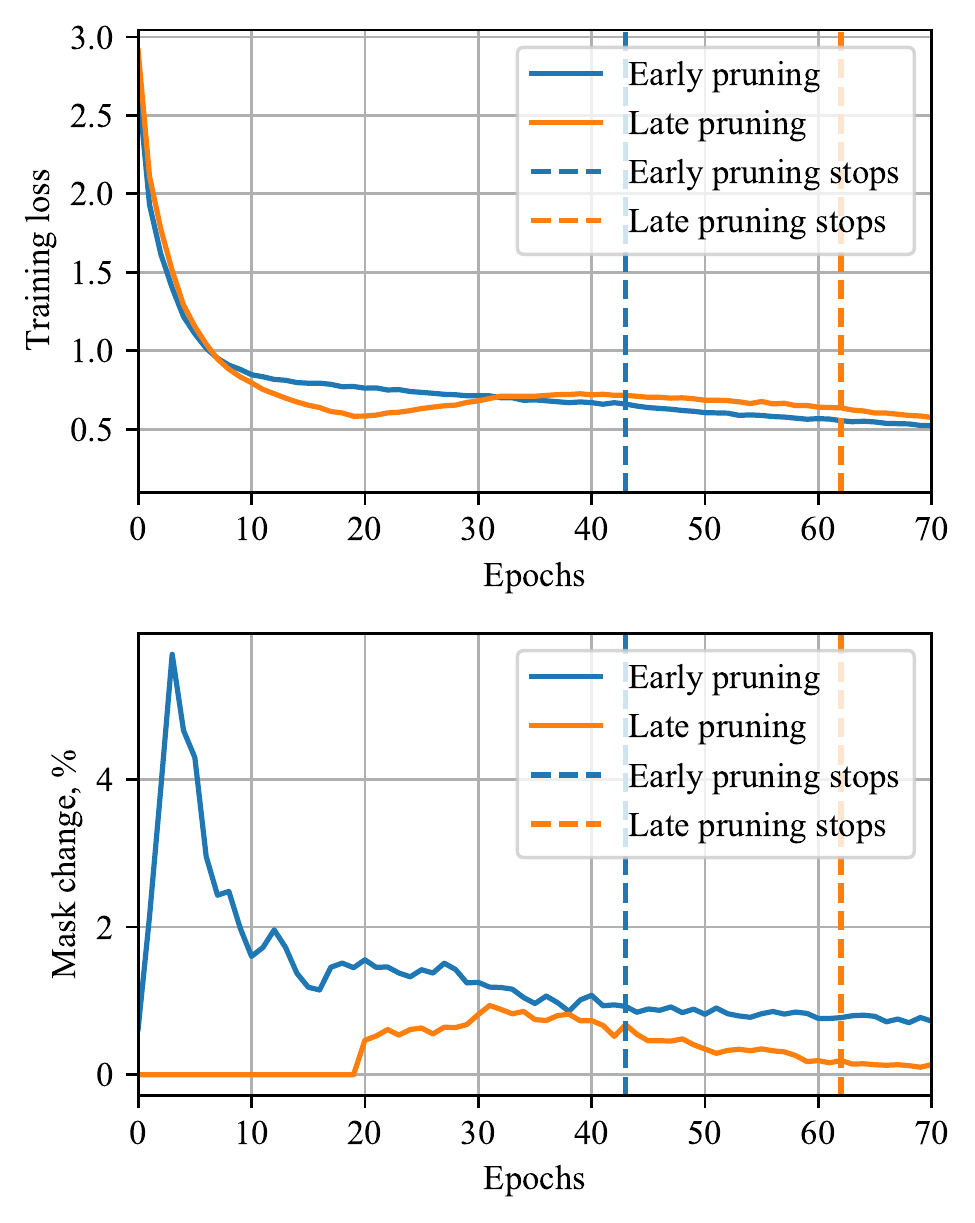}%
    \caption{The change in training loss (\emph{left}) and pruning masks, as a percentage of channels that were masked or unmasked from the last update (\emph{right}), during early and late pruning---starting from the 1\textsuperscript{st} and 20\textsuperscript{th} epoch, respectively, on VGG-16.}
    \label{fig:vgg16-loss-mask}
\end{figure}

To illustrate how the starting point of pruning affects the training of the network, Figure~\ref{fig:vgg16-loss-mask} plots the training loss and the percentage of channels that have changed their mask value since the previous update for VGG-16 architecture. The pruning is shown in two modes: starting at the 1\textsuperscript{st} epoch (early pruning) and 20\textsuperscript{th} epoch (late pruning).
The data shows that pruning eventually removes channels that contribute to generalisation, resulting in a spike in training loss (visible in late pruning yet smoothed over by rapid improvement in early training stages). Early pruning has more epochs remaining to recover from generalisation loss but has relatively unstable pruning masks (slow convergence) due to less knowledge of which layers are more important at the start (salience values are closer to uniform). 

From the training loss graph, one can conclude that early pruning is a safe choice of mode for pruning. However, if early termination is used, the final model performance may vary due to a sizeable fraction of mask change happening after the layer size learning has stopped. The sensitivity to the pruning learning rate is investigated in Appendix~\ref{apx:hyperparam-sensitivity}, which shows marginally better classification accuracy for lower learning rates (longer pruning). Overall, any performance loss or gain due to early termination will depend on a particular case of the architecture and compression rates requested, and is likely to be minor, as previously seen in the MobileNet-v2 example in Figure~\ref{fig:vww-mbnet-pru}.

\begin{figure}[ht!]
\centering
\begin{minipage}{0.45\textwidth}
\includegraphics[width=\textwidth]{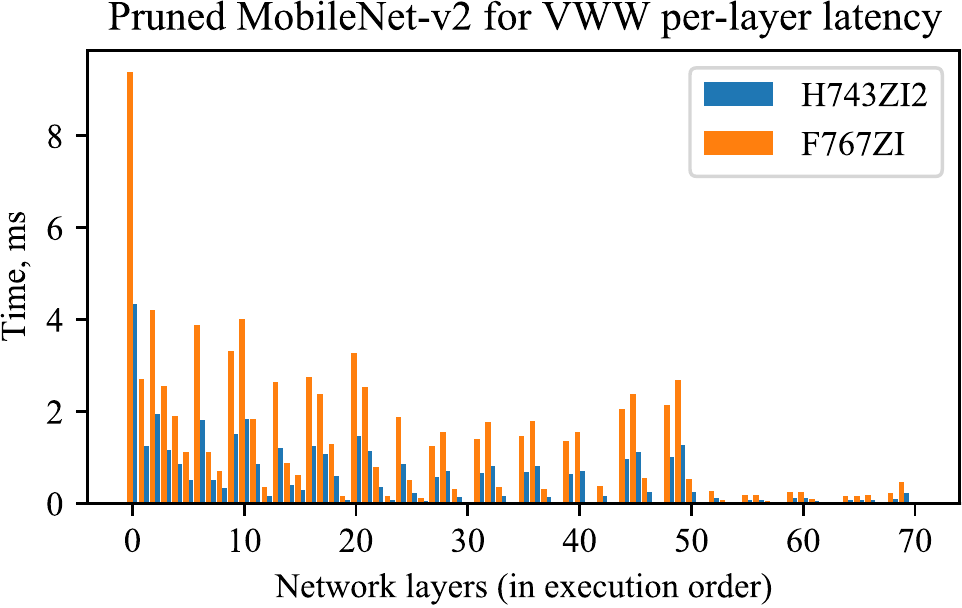}\\[3mm]
\includegraphics[width=\textwidth]{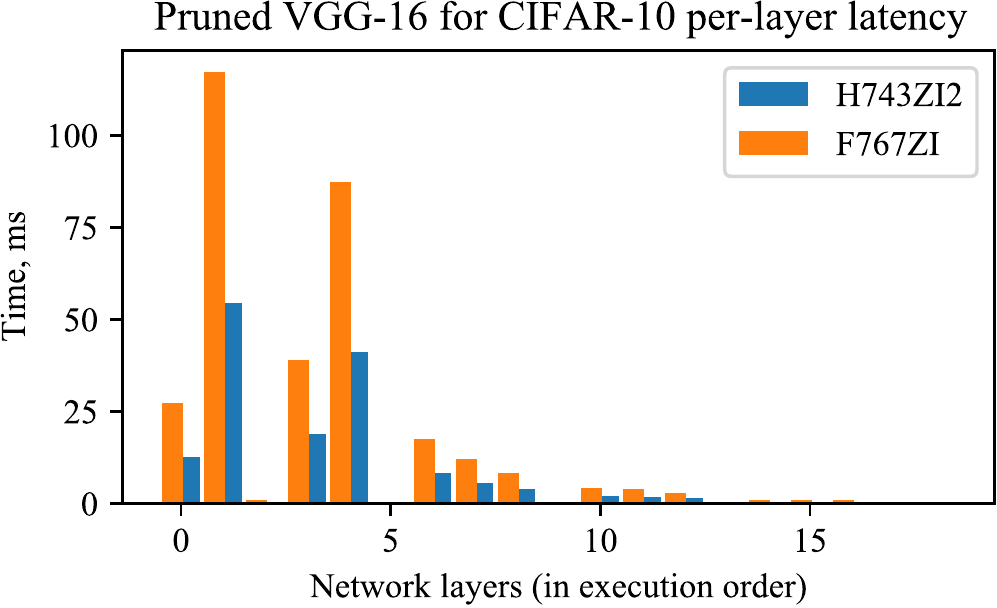}\\[3mm]
\includegraphics[width=\textwidth]{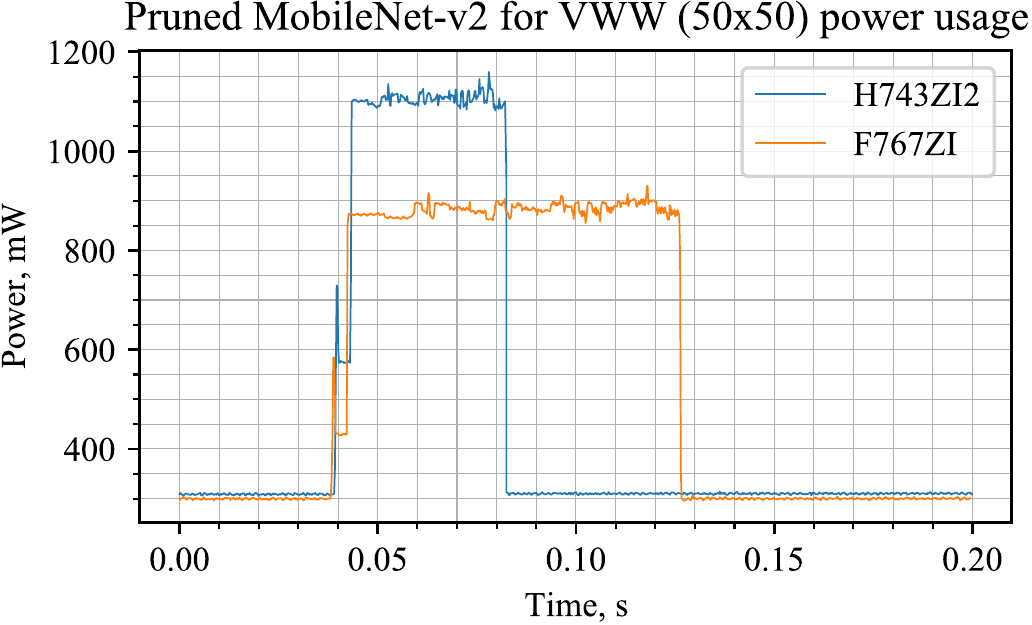}\\[3mm]
\includegraphics[width=\textwidth]{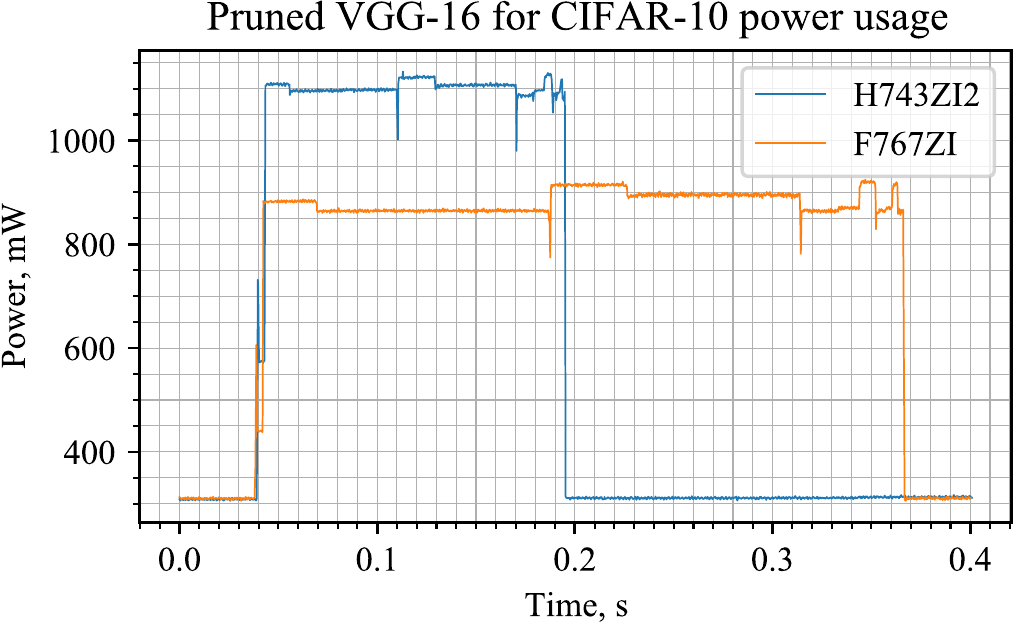}
\end{minipage}
\caption{Per-layer latency and power draw of the two analysed models--MobileNet v2 for VWW and VGG-16 for CIFAR-10---measured on Nucleo H743ZI2 and F767ZI development boards.}
\label{fig:latency-power-plot}
\end{figure}

\subsection{Deployment analysis}

Differentiable pruning makes a practical difference in deploying neural networks onto microcontroller-powered devices: only pruned versions of networks considered here would satisfy strict MCU computational constraints. To establish that pruned models have viable latency and power usage properties when executed on MCUs, the MobileNet-v2 and VGG-16 networks are deployed on two MCU development boards---\href{https://www.st.com/en/evaluation-tools/nucleo-f767zi.html}{NUCLEO-F767ZI}~\cite{nucleof767zi} and \href{https://www.st.com/en/evaluation-tools/nucleo-h743zi.html}{NUCLEO-H743ZI2}~\cite{nucleoh743zi2}---using the off-the-shelf TensorFlow Lite Micro~\cite{david2020tensorflow} runtime. Both boards are powered by the ARM Cortex-M7 chip, with clock speeds of 216 MHz and 480 MHz.

\begin{table}[h]
\centering
\small
\begin{tabular}{lrr}
\toprule
\textbf{Configuration} & \textbf{Latency, ms} & \textbf{Energy, mJ} \\
\midrule
\multicolumn{3}{l}{\textit{on Nucleo F767ZI:}} \\
MobileNet-v2 for VWW & 83.99 $\pm$ 0.01 & 75.8 $\pm$ 0.1 \\
VGG-16 for CIFAR-10 & 324.02 $\pm$ 0.03 & 287.5 $\pm$ 0.1 \\
\midrule
\multicolumn{3}{l}{\textit{on Nucleo H743ZI2:}} \\
MobileNet-v2 for VWW & 38.72 $\pm$ 0.01 & 45.6 $\pm$ 0.1 \\
VGG-16 for CIFAR-10 & 152.22 $\pm$ 0.04 & 169.7 $\pm$ 0.1 \\
\bottomrule
\end{tabular}
\caption{Mean latency and power usage for deployed configurations with 2 standard deviation errors from 30 measurements.}
\label{tab:ch5:depl-measurements}
\end{table}

Table~\ref{tab:ch5:depl-measurements} shows mean latency and energy usage for all four configurations. The pruned networks are viable for microcontroller hardware: both models have 10s--100s ms inference time and 10s--100s mJ energy usage per inference. 
For the same model, H743ZI2 executes over 2$\times$ faster due to the difference in clock speed; this is also confirmed by layer-wise latency plots presented in Figure~\ref{fig:latency-power-plot}. The figure also shows the power draw of a single inference round of both boards for each model: faster inference on H743ZI2 comes at an increased power draw (875 mW to 1100 mW), though the overall energy usage is lower. Power draw remains approximately constant throughout the model execution, confirming that energy usage is a function of model latency.

Overall, pruning was instrumental in enabling on-device deep learning inference and produced models have viable latency and energy usage to be used in real-world microcontroller-powered devices and systems. A variety of latency and energy usage requirements can be targeted by considering different model architectures or hardware.

\subsection{Application to ubiquitous computing} 
Considerable effort has been devoted to applying on-device deep learning to new domains, particularly in \emph{ubiquitous computing}, where microcontroller-powered wearable, embedded, and IoT devices are used for diverse tasks, such as fitness tracking, health monitoring, home security and voice assistance. Developing a successful deep learning solution for ubiquitous tasks requires domain knowledge of the task at hand, a neural network architecture that can capture the structure of the domain-specific input data and learn from it, and working around the computational limitations of microcontroller hardware. The development requires a wide range of competencies that a single developer or a small team may not possess, creating a high barrier to entry. 

On-device deep learning can be made more accessible to domain specialists by automating the resource usage reduction stage of the development process. Creating a general solution for automated model compression requires finding a compromise between modifying the architecture of the network yet largely preserving the expertise that went into its creation. This work bridges the gap between the resource footprint of well-performing neural network architectures and the resource scarcity of ubiquitous computing platforms by introducing \emph{a structured network pruning algorithm}, specialised for microcontroller hardware. 

Differentiable pruning has been evaluated on common low-footprint classification tasks for comparison purposes; however, these tasks have parallels to challenges faced by the real-world "smart" ubiquitous computing systems. For example, object classification datasets of different difficulties mirror person detection for the privacy of third parties~\cite{dimiccoli2018mitigating}; akin to keyword spotting (via spectral feature extraction), many applications apply CNNs to acoustic and motion sensor data, preprocessed using spectral analysis techniques, e.g. filter bank coefficients or MFCCs, which are generic and used across many tasks~\cite{radu2018multimodal}, such as human activity recognition~\cite{hossain2018deactive, xue2020deepmv}, object recognition~\cite{gong2019knocker} or handwriting recognition~\cite{yin2020learning} through acoustic sensing.
The evaluation has shown no sensitivity to the nature of the task, which suggests the methodology may scale beyond the considered tasks.

Overall, differentiable pruning is a worthwhile addition to the model compression toolkit for ubiquitous computing developers who want to leverage on-device deep learning. The methodology can produce models with comparable or improved resource usage to state-of-the-art solutions while offering fast compression and microcontroller specialisation. Domain experts can introduce differentiable pruning to their model development pipeline with negligible overhead, straightforward hyperparameter tuning (Appendix~\ref{apx:hyperparam-sensitivity}), and request models just by specifying available computational resources.

\section{Conclusion}

Differentiable network pruning enables on-device deep learning inference on extremely resource-constrained platforms, such as IoT devices. Through microcontroller specialisation and accurate resource budgeting within pruning, our methodology achieves high model compression rates and improves upon the model resource usage or compression speed, compared to prior work.










\section*{Acknowledgements}
This work was supported by Samsung AI and by the UK's Engineering and Physical Sciences Research Council (EPSRC) with grants EP\slash R018677\slash 1 (the OPERA project), EP\slash M50659X\slash 1 and EP\slash S001530\slash 1 (the MOA project) and the European Research Council via the REDIAL project (Grant Agreement ID: 805194).

\bibliography{paper}
\bibliographystyle{mlsys2022}


\newpage
\clearpage

\appendix

\section{Hyperparameter sensitivity}
\label{apx:hyperparam-sensitivity}


Differentiable pruning learns layer sizes $\vec{\pi}$ by optimising a linear combination of the task-specific and resource-specific losses, as an outer-level gradient descent optimisation process to network training. The full SGD update step to $\vec{\pi}$ is given as follows. Note the use of a clipping function (omitted from the Algorithm~\ref{alg:pruning-algo} for conciseness), which is used to clip gradient values to $[0, 0.025)$ for all experiments.
\begin{equation*}
    \vec{\pi}' = \vec{\pi} - \eta_\pi \: \text{clip}(\alpha_\textsc{RES} \frac{\partial \mathcal{P}_\textsc{RES}}{\partial \vec{\pi}} + \alpha_\textsc{TSK} \frac{\partial \mathcal{P}_\textsc{TSK}}{\partial \vec{\pi}})
\end{equation*}

From the equation, one can see that $\alpha_\textsc{TSK}$ and $\alpha_\textsc{RES}$ must be proportional to each other and bounded, for the following reasons:
\begin{itemize}
    \item If $\alpha_\textsc{RES} \gg \alpha_\textsc{TSK}$ (or $\alpha_\textsc{TSK} \approx 0$), the contribution of $\mathcal{P}_\textsc{TSK}$ is erased.
    \item If $\alpha_\textsc{RES} \ll \alpha_\textsc{TSK}$ (or $\alpha_\textsc{RES} \approx 0$), pruning fails to converge because $\frac{\partial \mathcal{P}_\textsc{TSK}}{\partial \vec{\pi}}$ is always negative and greater than $\frac{\partial \mathcal{P}_\textsc{RES}}{\partial \vec{\pi}}$, which prevents $\vec{\pi}$ from being reduced.
    \item If $\alpha_\textsc{RES} \gg 0$ and $\alpha_\textsc{TSK} \gg 0$, the sum of two gradients has high variance, resulting in a smaller useful range of values contained within the clipping bounds.
\end{itemize}

\begin{figure*}[h]
\centering
\includegraphics[width=0.45\textwidth]{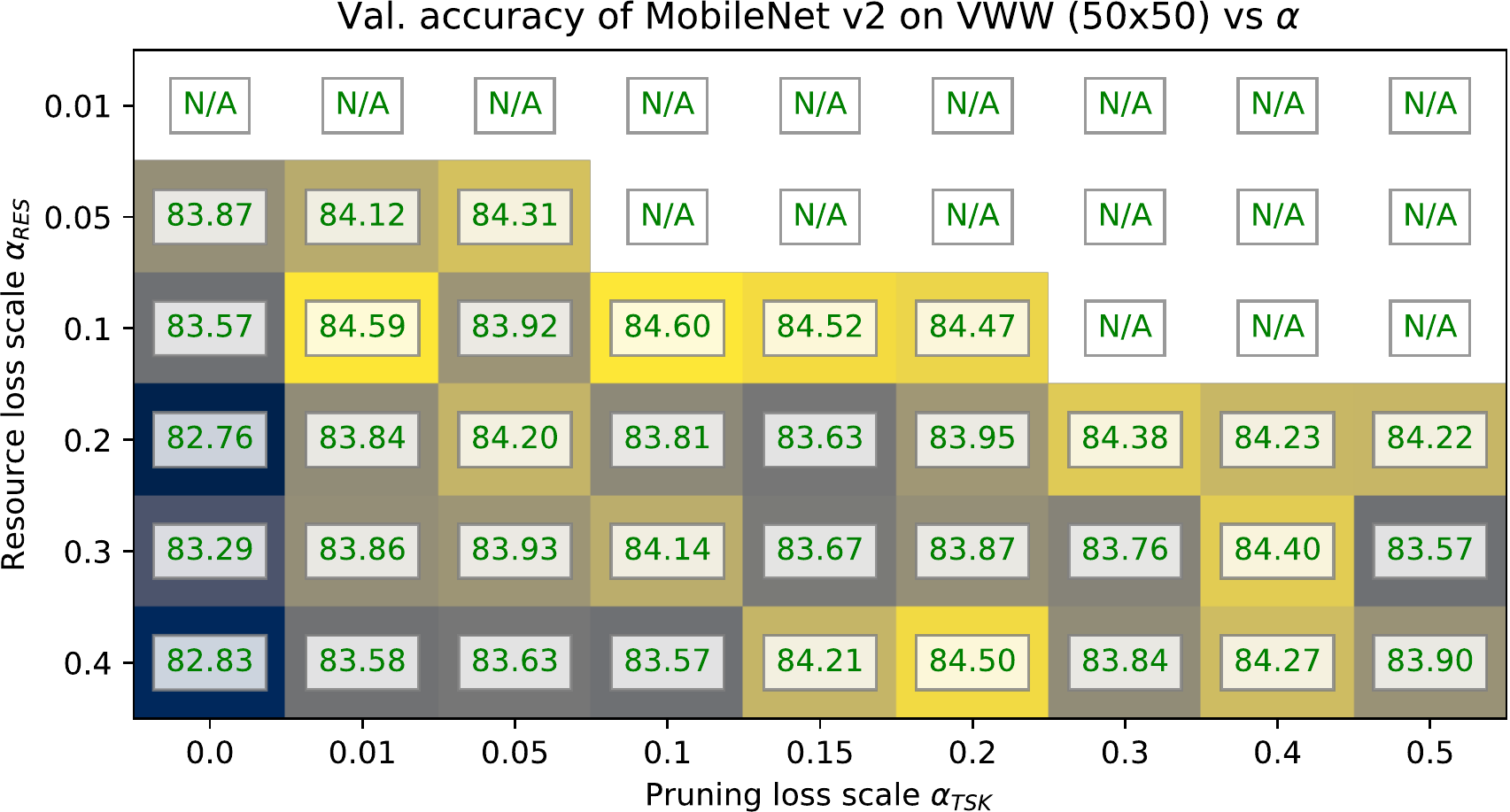}\quad
\includegraphics[width=0.45\textwidth]{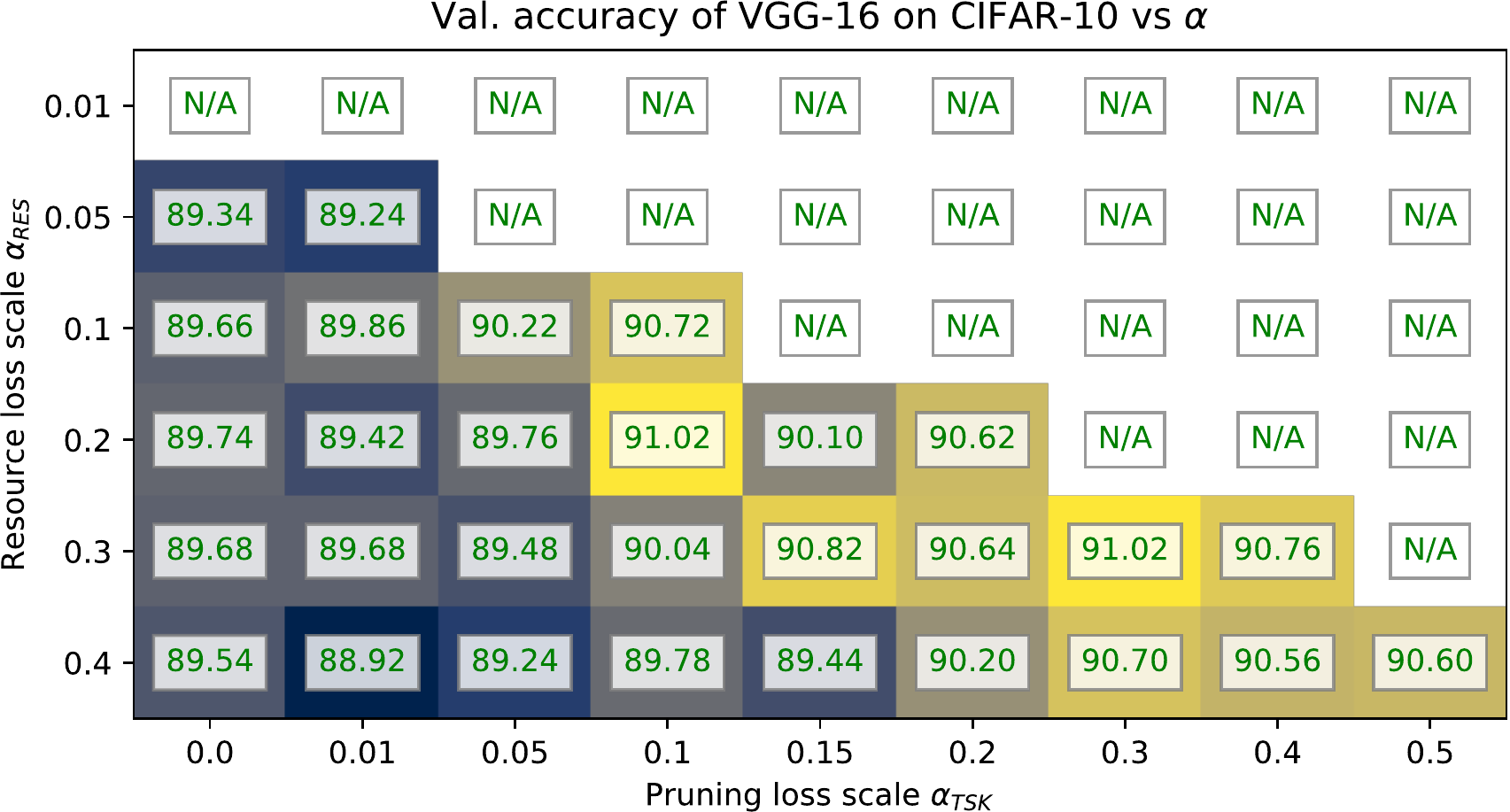}
\caption{Validation accuracy \emph{vs} $\alpha_\textsc{RES}$ and $\alpha_\textsc{TSK}$ for (\textit{left}) MobileNet-v2 on VWW (50$\times$50) and (\textit{right}) VGG-16 on CIFAR-10. Brighter is better; "N/A" refers to failure to converge before the network training completes.}
\label{fig:a02:hparam-alpha}
\end{figure*}

Figure~\ref{fig:a02:hparam-alpha} tabulates the validation accuracy \emph{vs} $\alpha_\textsc{RES}$ and $\alpha_\textsc{TSK}$ for MobileNet-v2 on VWW (50$\times$50 input, matching the resource usage of the MicroNets VWW-2 baseline) and VGG-16 on CIFAR-10 (matching the resource usage of the DSA baseline). For completeness, cases with $\alpha_\textsc{TSK} = 0$ are also included, corresponding to resource feedback-only pruning. The validation accuracy, however, has multiple sources of noise, such as initial weight values and quantisation errors. The results show the accuracy difference between the best and worst configurations of $\pm 1.8\%$ and $\pm 2.1\%$ for either model, respectively. The largest difference between adjacent cells (excluding $\alpha_\textsc{TSK} = 0$), which can be seen as a proxy for noise, is $\pm 0.65\%$ and $\pm 1.38\%$ for either model, respectively. 

Overall, the data exhibits a trend towards the largest possible values for $\alpha_\textsc{TSK}$ that do not quite prevent pruning from converging. However, a large fraction of the differences can be explained by noise, and regardless of the hyperparameter choice, the validation accuracy lies within a narrow range. This leads to the conclusion that the method has low sensitivity to hyperparameters $\alpha$.

How does one determine acceptable ranges of $\alpha$ to try? Luckily, there is no need to run multiple training iterations until completion. In practice, we found that starting with $\alpha_\textsc{RES}$ and $\alpha_\textsc{TSK}$ set such that the first $\frac{\partial \mathcal{P}_\textsc{RES}}{\partial \vec{\pi}}$ and $\frac{\partial \mathcal{P}_\textsc{TSK}}{\partial \vec{\pi}}$ lie within the same order of magnitude as the gradient cap ($0.025$) is sufficient---even if the hyperparameters are not optimal, the result will still be close to the best possible configuration.

\begin{figure*}[b]
    \centering
    \includegraphics[scale=0.7]{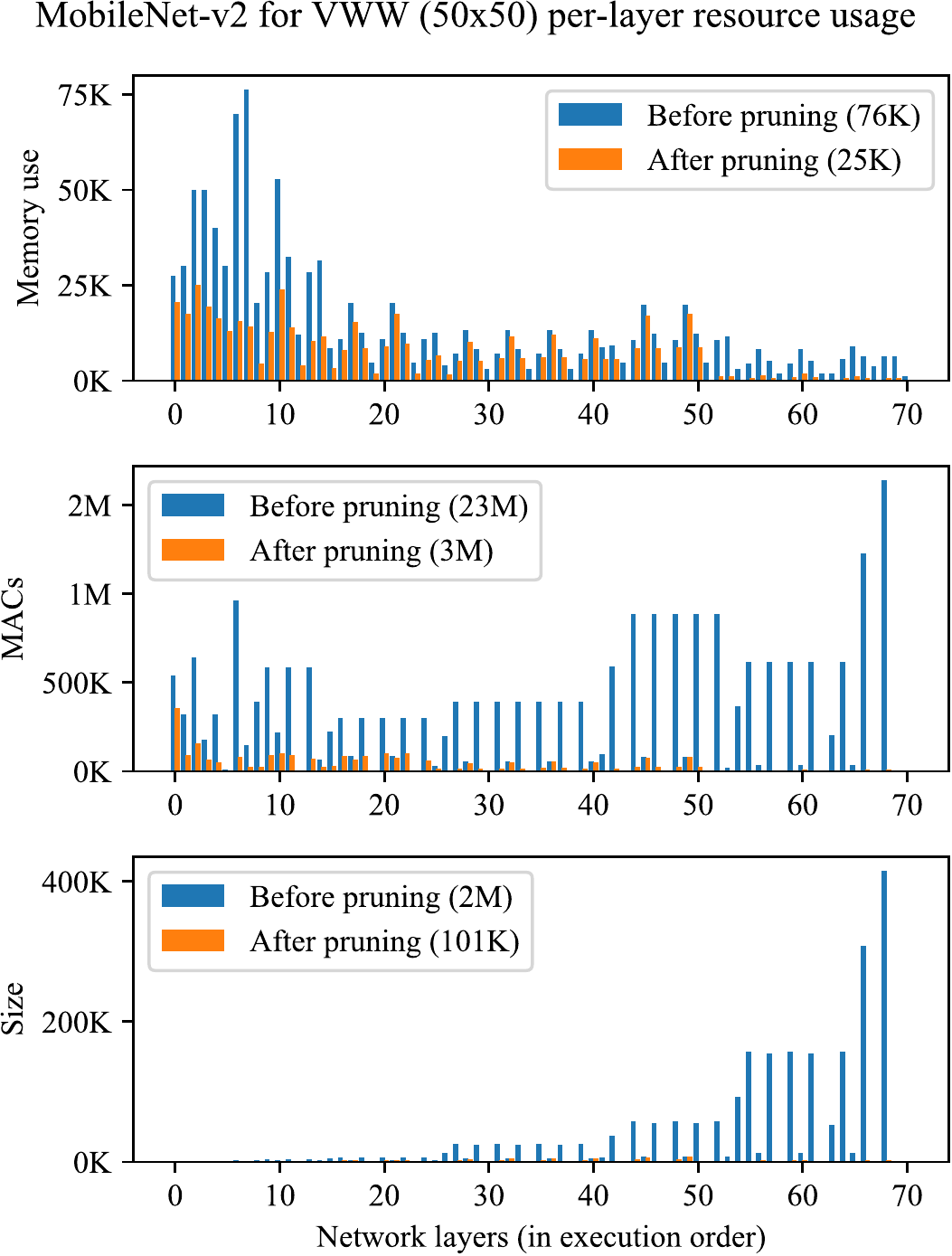}%
    \includegraphics[scale=0.7]{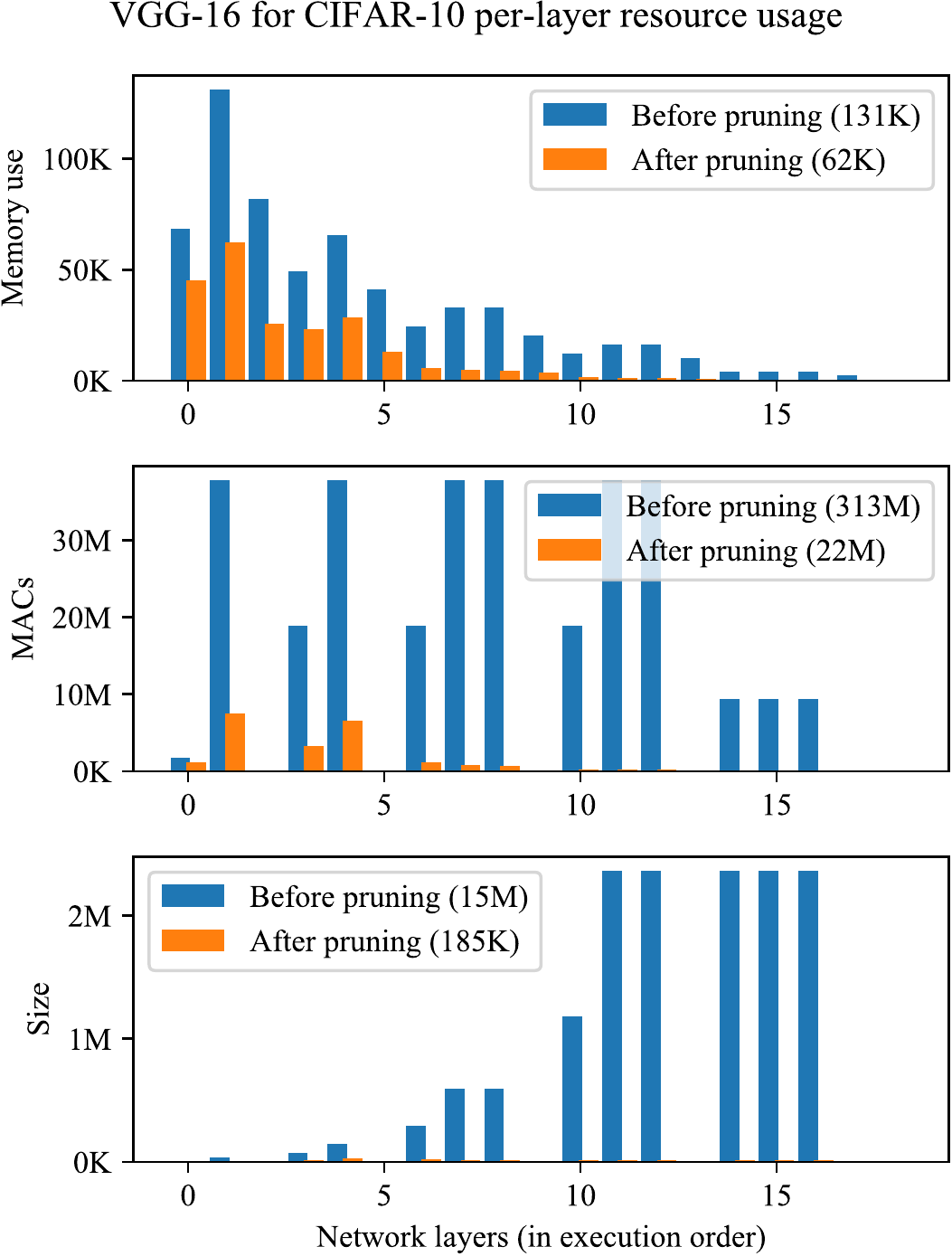}%
    \caption{Layer-wise resource usage before and after pruning for (\textit{left}) MobileNet-v2 on VWW (50$\times$50) and (\textit{right}) VGG-16 on CIFAR-10.}
    \label{fig:a02:res-plot}
\end{figure*}

\begin{figure*}[h]
    \centering
    \includegraphics[scale=0.45]{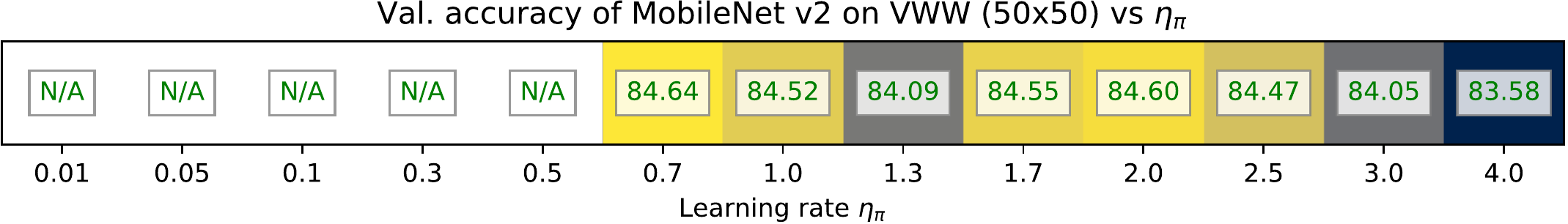}\\[2mm]
    \includegraphics[scale=0.45]{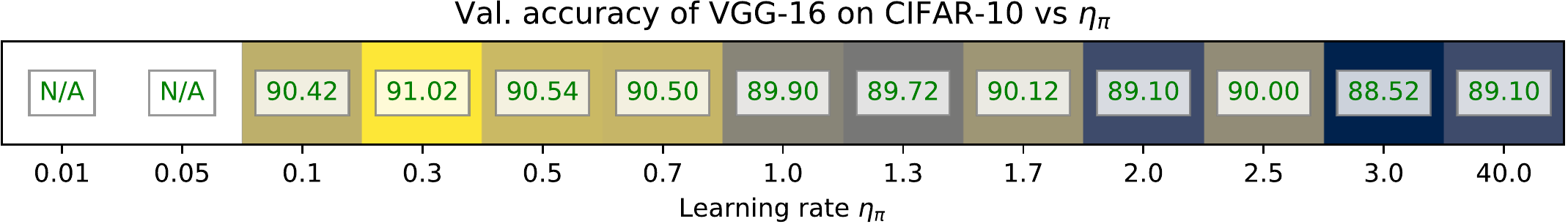}
    \caption{Validation accuracy \emph{vs} learning rate $\eta_\pi$ for (\textit{top}) MobileNet-v2 on VWW (50$\times$50) and (\textit{bottom}) VGG-16 on CIFAR-10. Brighter is better; "N/A" refers to failure to converge in time for training end.}
    \label{fig:a02:hparam-lr}
\end{figure*}
    
Figure~\ref{fig:a02:hparam-lr} shows the relationship between the validation accuracy and the learning rate $\eta_\pi$ for the same pair of models. When $\eta_\pi$ is too low, pruning fails to converge in time. The results show a wide range of near-optimal values for $\eta_\pi$, with the best-worst configuration difference of $\pm 1\%$ and $\pm 2.5\%$ for either model, respectively. Adjacent-cell value differences are within $\pm 0.5\%$ for both models. As with hyperparameters $\alpha$, the data shows a wide range of permissible values with a large fraction of accuracy differences explained by noise.

The results show a trend towards the lowest possible learning rates, which extend pruning throughout the entire training process. This results in the model reducing its capacity very gradually throughout training to meet the resource requirements, and ostensibly prevents the use of early termination (see Section~\ref{sec:ch5:discussion-early-term}). Therefore, a reduction in the training computation may come at the cost of classification accuracy.

\section{Post-pruning architecture analysis}
\label{apx:arch-anal}

\begin{figure}[p]
    \centering
    \includegraphics[scale=0.85]{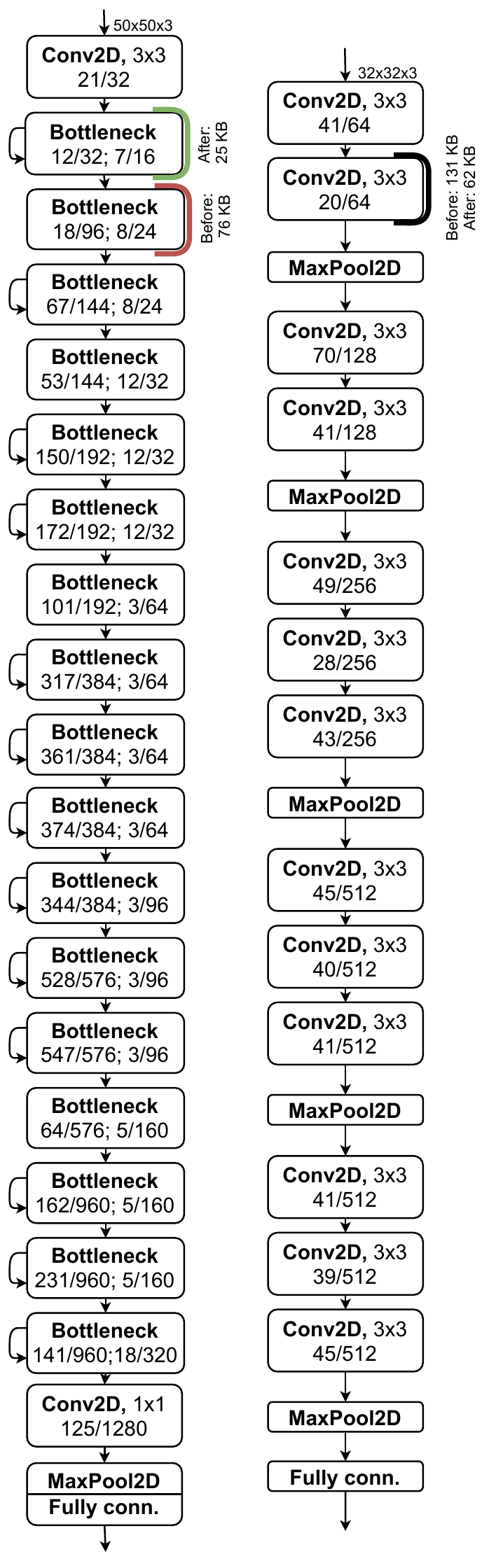}%
    \caption{MobileNet-v2 (\textit{left}) and VGG-16 (\textit{right}) architectures. Legend: "\textit{x/y}" refers to \textit{x} out of \textit{y} surviving channels after pruning; "\textbf{Bottleneck} \textit{a; b}" refers to an (inverted) bottleneck layer with \textit{a} bottleneck and \textit{b} output channels; self-arrow denotes a residual connection within. Brackets show the memory bottleneck before and after pruning.}
    \label{fig:a02:pruned-arch}
\end{figure}

To gain more insight into how differentiable pruning changes the network architecture and resource usage distribution across layers, two models are analysed: MobileNet-v2 trained on the VWW (50x50) dataset and VGG-16 trained on the CIFAR-10 dataset. 

Figure~\ref{fig:a02:pruned-arch} (page~\pageref{fig:a02:pruned-arch}) shows layer sizes and memory bottlenecks of both architectures before and after pruning. The results show that: (a) pruning does target the memory bottleneck, reducing it by over 2$\times$ for both models; (b) in MobileNet-v2, layers on the main feature extraction path (output layers of the bottleneck blocks, as well as all layers in the bottleneck blocks without a residual connection) have been pruned more to reduce the memory usage; (c) in VGG-16, the pruning has undone the increment in channel dimension as depth increases, resulting in a model that is approximately uniformly wide across all layers.

Figure~\ref{fig:a02:res-plot} shows per-layer resource usage before and after pruning: memory usage (working set size for each layer), the number of MACs and parameter size. The tails of convolutional architectures have a small spatial resolution and a high number of channels, which results in a large increase in size and greater model capacity. In both instances, pruning has effectively performed depth adjustment of the models by pruning more in the tail end of the model (most easily seen in the ``Memory use'' graph).

Through manual inspection of pruned network architectures, differentiable pruning is confirmed to target microcontroller resource bottlenecks, in a way that is consistent with removing extraneous complexity from the model.


\end{document}